\pdfoutput=1
\documentclass[11pt]{article}

\usepackage{amsfonts, amsmath, amssymb, amsthm}
\usepackage[dvipsnames]{xcolor}
\usepackage{mathtools}
\usepackage[shortlabels]{enumitem}
\usepackage{hyperref}
\usepackage{booktabs}
\usepackage{paralist}
\usepackage{longtable}
\usepackage{subcaption} 
\usepackage{tikz}
\usepackage[boxed]{algorithm}
\usepackage[noend]{algpseudocode}
\allowdisplaybreaks
\tikzstyle{vertex}=[circle, draw, inner sep=0pt, minimum size=6pt]
\usepackage{xspace}
\usepackage[]{acl}

\usepackage{times}
\usepackage{latexsym}

\usepackage[T1]{fontenc}
\usepackage[utf8]{inputenc}

\usepackage{microtype}

\usepackage{inconsolata}

\author{Sayan Ghosh \quad Tejas Srinivasan \quad Swabha Swayamdipta \\
Thomas Lord Department of Computer Science, University of Southern California\\
\texttt{\{ghoshsay, tejas.srinivasan, swabhas\}@usc.edu} 
}

\newcommand{\comm}[1]{}

\newcommand{\mycomment}[1]{}

\newcommand{\draftonly}[1]{#1}
\renewcommand{\draftonly}[1]{}

\newcommand{\separability}{\textsc{separability}\xspace}
\newcommand{\gpt}{\texttt{GPT-3.5}\xspace}
\newcommand{\vicuna}{\texttt{Vicuna-7B}\xspace}
\newcommand{\mistral}{\texttt{Mistral-7B}\xspace}
\newcommand{\flan}{\texttt{FLAN-T5-XXL}\xspace}
\newcommand{\ELO}{\textsc{ELO}\xspace}
\newcommand{\sepELO}{\textsc{sep-ELO}\xspace}

\title{Compare without Despair: \\Reliable Preference Evaluation with Generation \separability}
\begin{document}

\maketitle

\widowpenalty10000

\begin{abstract}
Human evaluation of generated language through pairwise preference judgments is pervasive.
However, under common scenarios, such as when generations from a model pair are very similar, or when stochastic decoding results in large variations in generations, it results in \emph{inconsistent} preference ratings.
We address these challenges by introducing a meta-evaluation measure, \separability, which estimates how suitable a test instance is for pairwise preference evaluation. 
For a candidate test instance, \separability samples multiple generations from a pair of models, and measures how \emph{distinguishable} the two sets of generations are. 
Our experiments show that instances with high \separability values yield more consistent preference ratings from both human- and auto-raters. 
Further, the distribution of \separability allows insights into which test benchmarks are more valuable for comparing models. 
Finally, we incorporate \separability into \ELO ratings, accounting for how suitable each test instance might be for reliably ranking LLMs.  
Overall, \separability has implications for consistent, efficient and robust preference evaluation of LLMs with both human- and auto-raters.
\end{abstract}

\section{Introduction}
\label{sec:intro}

As large language models' (LLM's) capabilities have rapidly improved in recent years, evaluation of these capabilities has become increasingly reliant on human preference judgments that compare pairs of model generations.
While these judgments offer freedom from gold-standard references \citep{papineni-etal-2002-bleu,lin-2004-rouge,zhangbertscore}, they are far from perfect \cite{gehrmann2022repairing}.\looseness=-1

In particular, human evaluation faces issues including low rater agreements \cite{goyal2022news}, spurious correlations with factors like length \citep{wu2023style, sun-etal-2019-compare}, lack of measurement validity~\citep{ethayarajh2022authenticity}, and inconsistent usage and interpretation of inter-rater agreement \citep{amidei-etal-2019-agreement, prabhakaran-etal-2021-releasing}.
Furthermore, for annotation efficiency, human judgments are sometimes replaced with LLM judgments, which have shown high correlation with crowdworker ratings \citep{dubois2024length, zheng2024judging, lin2024wildbench, liu-etal-2023-g, zeng2023evaluating}; however, it remains unclear whether such auto-evaluations are a step in the right direction or exacerbate existing biases \citep{zheng2024judging, wang2023large, wu2023style, chang2024survey}.\looseness=-1

\begin{figure*}[ht!]
  \centering
    \includegraphics[width=\textwidth]{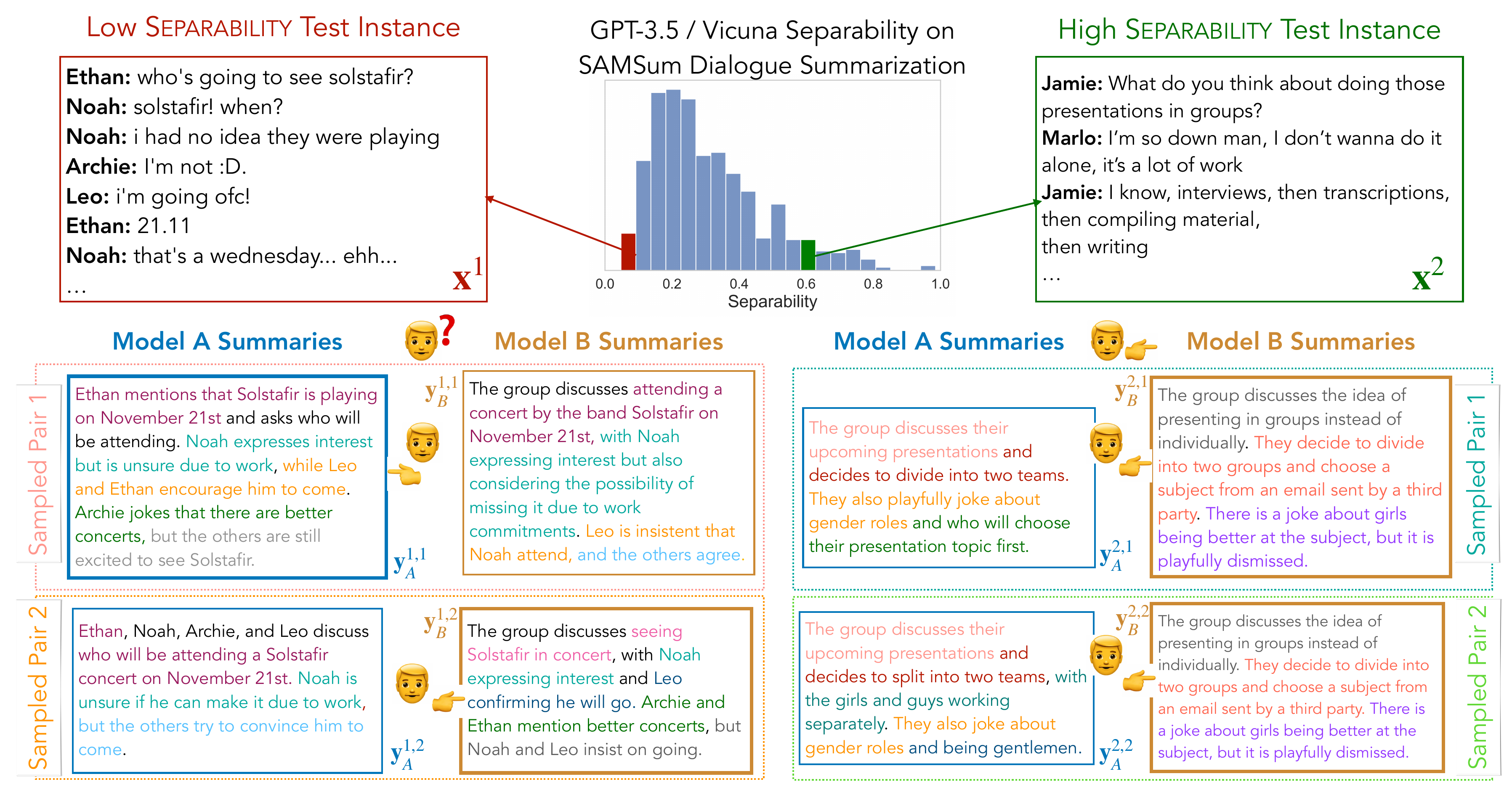}
  \caption{Illustration of \separability on SAMSum dialog summarization from our human experiments (\S\ref{sec:human-study}). 
  Test instances have varying degrees of \separability, which lead to different levels of consistency in preference ratings.
  For lower \separability\ instances, the choice of which pair of sampled generations to show raters affects human rating (raters preferred Model A under Pair 1 and B under Pair 2); hence the overall judgment between model pairs is inconsistent. 
  Human preferences are consistent under higher \separability (raters always preferred Model B).
  \label{fig:sep-output-example}
  }
\end{figure*}
In this work, we focus on the problem of \emph{unreliable} preference judgments from human raters, illustrated in \autoref{fig:sep-output-example}.
We show that output pairs from any two modern LLMs can often be hard to distinguish from each other; such high \textbf{cross-alignment} between models can cause preference judgments to be highly arbitrary.
We identify another understudied factor affecting judgments: the \textit{variability within one LLM's generations for the same input}, owing to the stochasticity of popular decoding techniques such as temperature sampling \citep{giulianelli-etal-2023-comes, tsvilodub2024predictions}.
Such low \textbf{self-alignment}, in addition to high cross-alignment between models, may lead to \emph{inconsistent} ratings---highly dependent on the exact sampled pair chosen for preference judgments.
As a concrete example, on 100 news articles from CNN/DailyMail \citep{nallapati-etal-2016-abstractive}, our human evaluations show that when comparing five different summary pairs for each input, raters picked the same model only 46\% of the time (\S\ref{sec:human-study}).   
These findings raise the question: when can we rely on pairwise judgments to compare generations from two LLMs?

We argue that some test instances might be better suited for human evaluation than others, mirroring insights from prior work \cite{rodriguez-etal-2021-evaluation, vania-etal-2021-comparing}.
We introduce \textbf{\separability}, a meta-evaluation measure that determines, for a single instance, how distinguishable two sets of generations from two models are (\S\ref{sec:separability}). 
\separability builds on the intuition that the harder it is to distinguish generations from two models, the less consistent the preference ratings will be.
Our formulation of \separability combines cross-alignment between generations between pairs of models as well as self-alignment between multiple generations from each given model.
We operationalize self- and cross-alignment in \separability\ with a flexible choice of similarity metric depending on the salience of the variability (e.g. lexical, semantic) in the preference judgment.

Our experiments with \separability on different model pairs, benchmarks and tasks show that \separability\ can not only identify test instances which are likely to yield consistent preference ratings, but also benchmarks likely to yield consistent comparisons between models. 
For instance, we show that evaluation sets such as CNN/DailyMail \citep{nallapati-etal-2016-abstractive} are not as useful in comparing modern LLMs as they were in comparing earlier summarization-specific models, supporting prior findings \citep{goyal2022news, zhang2023benchmarking}.
Through extensive human evaluation, we show that instances with high \separability\ scores tend to result in more consistent preferences (\S\ref{sec:human-study}).
Moreover, we find that LLM-based auto-evaluation systems \cite{dubois2024length} also have similar patterns of consistency and inconsistency as human raters.\looseness=-1 

Finally, as a direct application of \separability, we extend it to \ELO, a rating system based on pairwise comparisons, now used widely for LLM generations  \citep{chiang2024chatbot}. 
By modifying the \ELO rating update function to account for the \separability of each new instance,
our \separability-ELO ratings provide more nuanced comparisons (\S\ref{sec:applying-sep}).
Overall, \separability\ offers a reliable signal in the noisy landscape of generative evaluation via pairwise ratings, provides insights into benchmarks and test instances, and complements existing ranking measures for robust preference evaluation.
Our code and data is available at \url{https://github.com/dill-lab/separability}.   

\section{\separability as a Meta-Evaluation}
\label{sec:separability}

We address the problem of consistency in modern generative evaluation: namely, how suitable a test input $\mathbf{x}_i \in \mathcal{X}$ is for collecting reliable preference ratings between output generations from a pair of models, $m_A$ and $m_B$.
Our approach is based on the key intuition that it is harder to collect consistent preference ratings between $m_A$ and $m_B$ if their output generations are, on average, harder to distinguish for a (human) rater.
For instance, the distinction is hard when the generations focus on similar content (see summaries in \autoref{fig:sep-output-example}, left), or have similar styles; we call this \textbf{high cross-alignment} between generations from $m_A$ and $m_B$.
Another factor that may make distinguishing two models' generations harder is large variability within each model's sampled generations, due to stochastic LLM decoding approaches such as temperature and nucleus sampling. 
Such variability, which we refer to as \textbf{low self-alignment}, makes it hard to characterize each model's specific tendencies, which in turn makes it hard to have a consistent preference for a single model. 
Under the above two conditions, the choice of which generations to use for pairwise comparison influences the preference rating outcome (\autoref{fig:sep-output-example}).\looseness=-1

Both kinds of alignments, while orthogonal, play a key role in determining how consistent human ratings for an instance might be (\S\ref{sec:calc-alignments}).
We introduce \textbf{\separability}, a meta-evaluation measure that estimates how suitable a test instance is for preference rating by consolidating cross- and self-alignment (\S\ref{sec:calc-sep}).
While \separability does not determine which generation is better or more preferable, it helps us understand how much we can trust each preference rating for a given input instance.

\begin{figure}[t!]
  \centering
    \includegraphics[width=\linewidth]{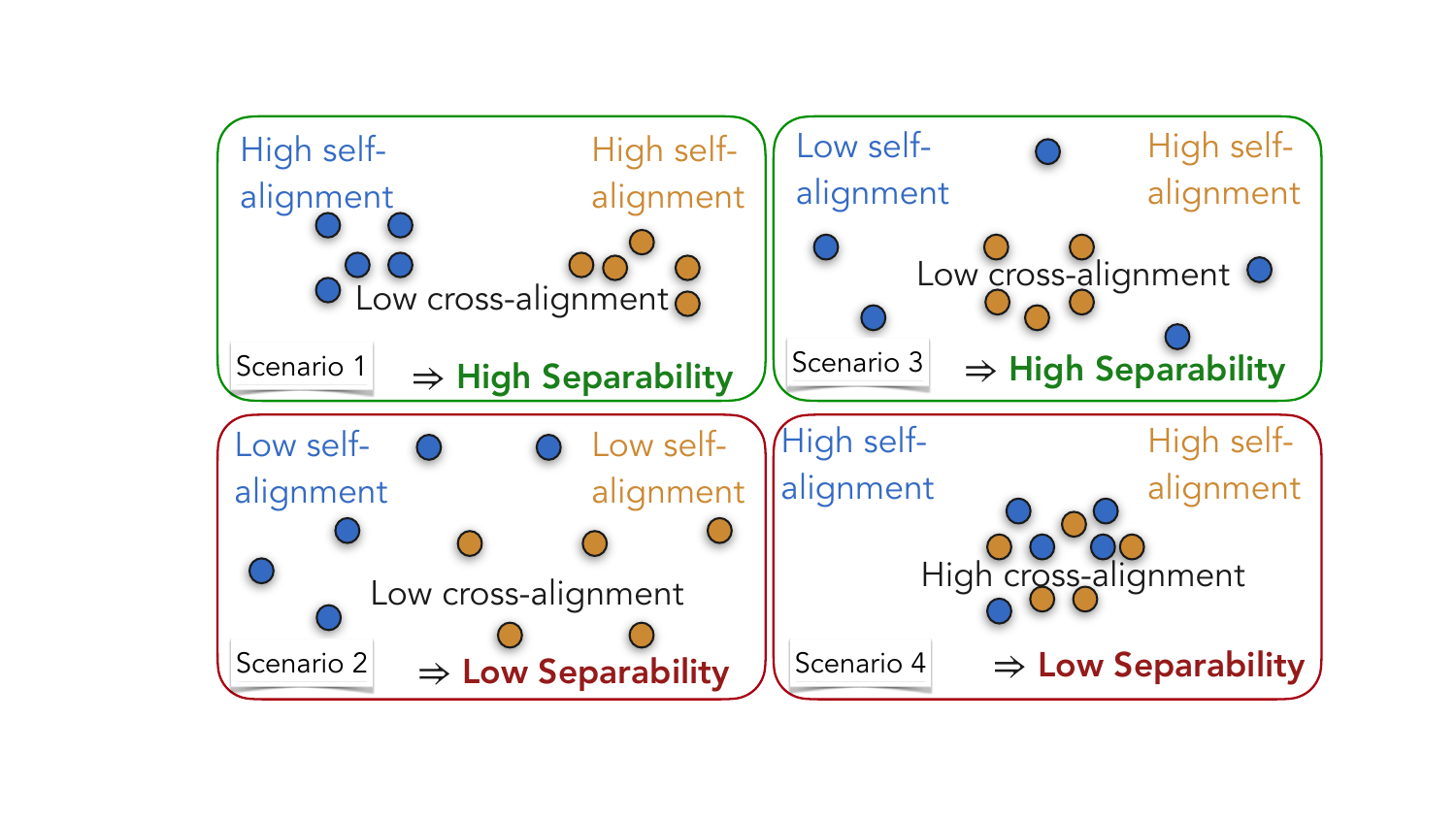}
    \caption{
    Four scenarios illustrating the intuition behind \separability. 
    Blue and gold circles represent generations from models $m_A$ and $m_B$ respectively, and Euclidean distances represent (dis)similarities between them.
    For a given input, at least one of the two models needs to have higher similarity among its own generations (high self-alignment) to have high \separability\ for that input.
    High similarity across generations from different models (high cross-alignment) leads to lower \separability.
    High self-alignment corresponds to low spread of a set of same-colored circles and vice-versa.
    High cross-alignment corresponds to low spread of the entire set of circles and vice-versa.
  }
  \label{fig:sep-intuition}
\end{figure}

At a very high level, there are four common scenarios which may occur in comparing generations from two models, which we highlight in \autoref{fig:sep-intuition}.\footnote{While these scenarios aren't fully exhaustive, they represent 97.5 \% of the roughly 30,000 instances that are used for experiments in this section.}
Scenarios 1, 2 and 3 all depict output sets where any sample from model $A$ is expected to be very distant from any sample from model $B$ (i.e. low \emph{cross-alignment}). 
It is easy to distinguish the two models under scenarios 1 and 3.
In Scenario 1, if two generations are very different, they must be from different models. 
For Scenario 3, the difference in self-alignments is a clear distinguishing factor between the two sets of generations.
Under scenario 2, generations \emph{from the same model} are also far from each other (i.e. the \emph{self-alignment} is also low), which makes the overall output sets hard to distinguish. 
In contrast, scenario 4 depicts a situation where both self- and cross-alignments are high; all generations,  regardless of the model they came from, are similar, making it hard for the rater to distinguish the models' output sets. 

Formally, given a set of test inputs $\{ \mathbf{x}^{i} \}_{i=1}^D$ (e.g.  news articles and instructions to summarize them), LLMs $m_A$ and  $m_B$ each induce a conditional distribution $p_{m_A}(\mathbf{y}^i \mid \mathbf{x}^i), p_{m_B}(\mathbf{y}^i \mid \mathbf{x}^i)$ over output generations $\mathbf{y}^i \in \mathcal{Y}$ (e.g. summaries of $\mathbf{x}^i$). 
From this distribution, we can sample $K$ generations using a common stochastic decoding algorithm such as temperature sampling, yielding sets $\{ \hat{\mathbf{y}}_A^{i, j}\}_{j=1}^K$ and $\{ \hat{\mathbf{y}}_B^{i, l}\}_{l=1}^K$.\looseness=-1

\subsection{Calculating Generation Alignments}
\label{sec:calc-alignments}

We define an \textbf{\emph{alignment function}}, $\mathcal{A}^i_{A, B}$ that estimates the similarity of two output distributions $p_{m_A}(\mathbf{y}^i \mid \mathbf{x}^i)$, $p_{m_B}(\mathbf{y}^i \mid \mathbf{x}^i)$ produced by LLMs $m_A, m_B$ on an input $\mathbf{x}^i$, 
\begin{equation}
    \label{eq:self-alignment}
    \mathcal{A}^i_{A, B} := %\\
    \mathbb{E}_{\mathbf{y}_A^i \sim p_{m_A}, \mathbf{y}_B^i \sim p_{m_B}} [s(\mathbf{y}_A^i, \mathbf{y}_B^i)],
\end{equation}
where $s: \mathcal{Y} \times \mathcal{Y} \to \mathbb{R}$ is a text similarity metric such as ROUGE \citep{lin-2004-rouge}, BERTScore \citep{zhangbertscore}, or BLEU \citep{papineni-etal-2002-bleu}. 
Intuitively, the alignment score $\mathcal{A}_{A, B}^i$ measures the expected similarity between an output from $m_A$ and an output from $m_B$, parameterized by $s$.
A high value of $\mathcal{A}_{A, B}^i$ indicates high similarity (low variability) between the generations of the two models.
Different similarity metrics can be used for different tasks.
In cases where a user cares about fine-grained lexical differences, a metric such as BLEU may be suitable.
On the other hand, if fine-grained lexical differences are less important than coarser semantic differences, metrics such as BERTScore or word embedding cosine similarity would be more suitable. 
We use a variation of BERTScore with a length-adjustment (defined in Section~\ref{sec:sim-fcns}), unless otherwise noted.

Since the space of model generations is intractable to calculating Equation~\ref{eq:self-alignment} exactly, in practice we use Monte-Carlo samples to approximate the alignment score. 
We sample $K$ generations from each model, resulting in output sets $\{ \hat{\mathbf{y}}_A^{i, j}\}_{j=1}^K$ and $\{ \hat{\mathbf{y}}_B^{i, l}\}_{l=1}^K$.
We approximate $\mathcal{A}_{A, B}^i$ as:
\begin{equation}
\label{eq:alignment-approx}
    \hat{A}_{A, B}^i = \frac{1}{K^2} \sum_{j=1}^K \sum_{l=1}^K s\left( \hat{\mathbf{y}}_A^{i, j}, \hat{\mathbf{y}}_B^{i, l}\right)
\end{equation}
When measuring \textbf{self-alignment}\footnote{For self-alignment, we skip $j = l$ terms in the summation.}---the level of variability in an individual model's output---we set $A=B$ in Equation \ref{eq:alignment-approx}.
When evaluating the variability between two distinct model, i.e. $A \neq B$, we label this function \textbf{\emph{cross-alignment}}.\footnote{We generate $K=5$ samples using temperature sampling with $\tau = 0.5$ as the default in our experiments; this corresponds to $K^2=25$ cross-alignment comparisons.}

\subsection{Calculating Generation \separability}
\label{sec:calc-sep}

Intuitively, in order to determine how distinguishable two models are, we need to measure the difference between the variability within each model's generation sets and the variability of the combined set of generations (i.e. the difference  between  each self-alignment and the cross-alignment). 
If the combined set has more variability than the variability within either model's generations, we consider the two generation sets to be \emph{separable}.

We define the generation \separability between models $A$ and $B$ for instance $i$, $\delta_{A,B}^i$ as:
\begin{equation}
\label{eq:separability}
\delta_{A,B}^i = \\
\max \left( \mathcal{A}_{A, A}^i, \mathcal{A}_{B, B}^i \right)
- \mathcal{A}_{A, B}^i.
\end{equation}
In \autoref{fig:sep-intuition}, under scenarios 1 and 3, generations of at least one model have low variability (and therefore high self-alignment); this combined with the low cross-alignment leads to higher \separability than in scenarios 2 and 4.\looseness=-1

\separability can take values in $[-1, 1]$.\footnote{We apply min-max normalization to constrain the range of alignments to $[0, 1]$ and to make alignment scores more interpretable and comparable across different model classes.}
In practice, however, cross-alignment is usually lower than self-alignment, making  $\delta_{A,B}^i \in [0, 1]$.\looseness=-1

\paragraph{Similarity Functions}
\label{sec:sim-fcns}
As our default similarity function $s$, we use a length-adjusted version of BERTScore \citep{zhangbertscore}, using the same length penalty used in BLEU \cite{papineni-etal-2002-bleu}.\footnote{$\textsf{LP} = \exp\left(1 - \frac{|\mathbf{y}^i_{\text{ref}}|}{|\mathbf{y}^i_{\text{hyp}}|}\right)$, where we fix $\mathbf{y}^i_{\text{ref}}$ to be the longer generation in order to ensure the penalty is symmetric}
In the case of translation, where more fine-grained lexical differences are important, we use BLEU itself. 
\autoref{appendix:alt-metrics} reports results with additional similarity functions: ROUGE-1 F1 \citep{lin-2004-rouge}\footnote{Using the \href{https://pypi.org/project/rouge-score/}{\texttt{rouge-score}} package.}, BLEU \citep{papineni-etal-2002-bleu}, Entity Similarity, and Cosine Similarity of DistilRoBERTa sentence embeddings.\footnote{Using \texttt{all-distilroberta-v1} in the \texttt{sentence-transformers} library} 
Due to the large variance in the range of each of these functions, we apply min-max normalization over the alignment values to constrain them to the range $[0, 1]$. 

Some of these metrics (e.g. BLEU) were designed to compare a ``candidate'' generation to a ``reference,'' we do not make this distinction since we do not use any reference generations. 
Instead, we arbitrarily choose the longer generation as the reference. 
We use F1-score variants of these metrics rather than recall or precision-oriented variants.
While prior work \cite{gehrmann2022repairing} shows that some of these similarity metrics are not optimal for reference-based \emph{evaluation}, we use these as textual \emph{similarity} functions.

\begin{figure*}[ht!]
     \begin{subfigure}[b]{0.49\textwidth}
         \centering
         \includegraphics[width=\textwidth]{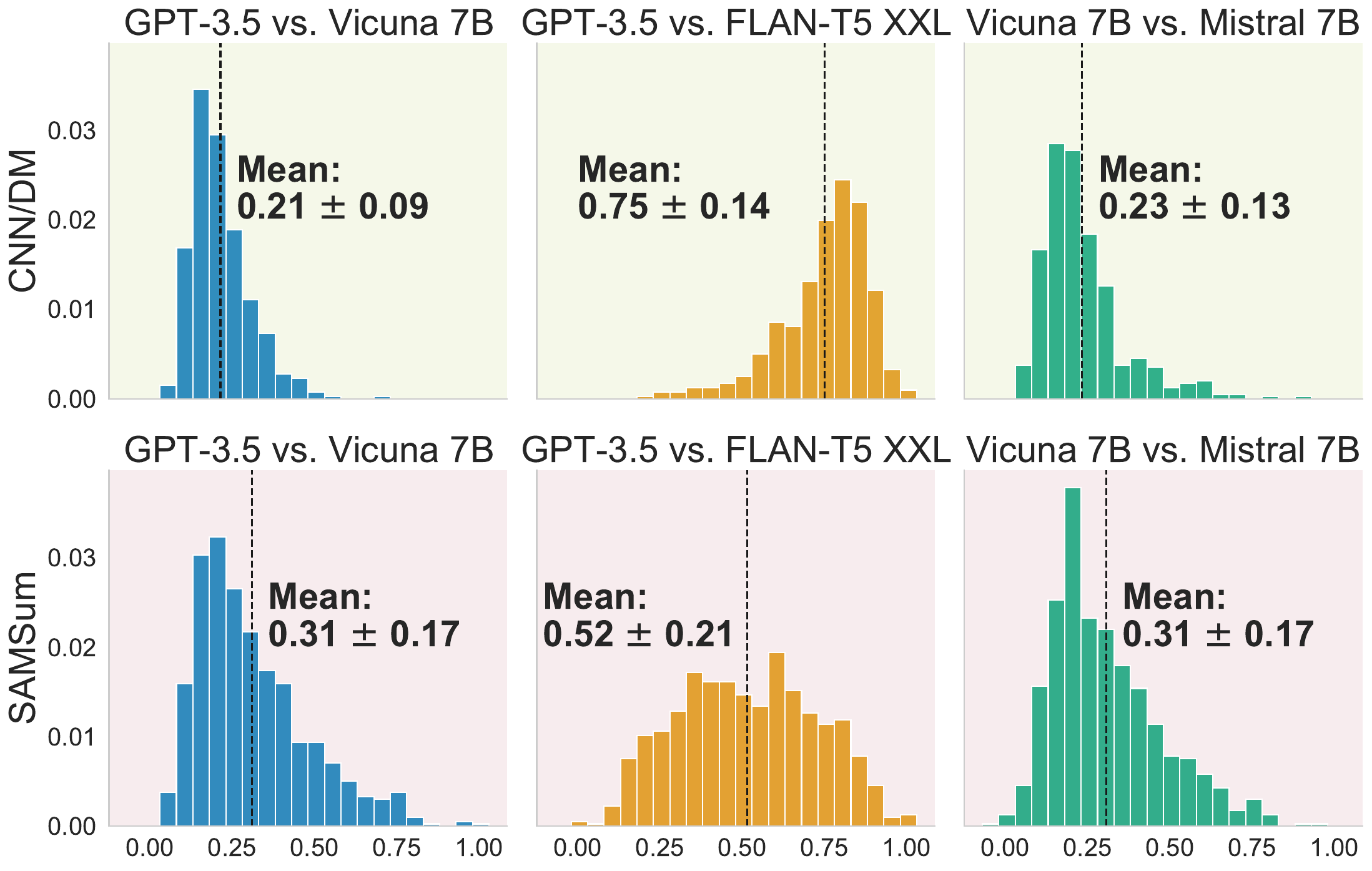}
     \end{subfigure}
     \begin{subfigure}[b]{0.49\textwidth}
         \centering
         \includegraphics[width=\textwidth]{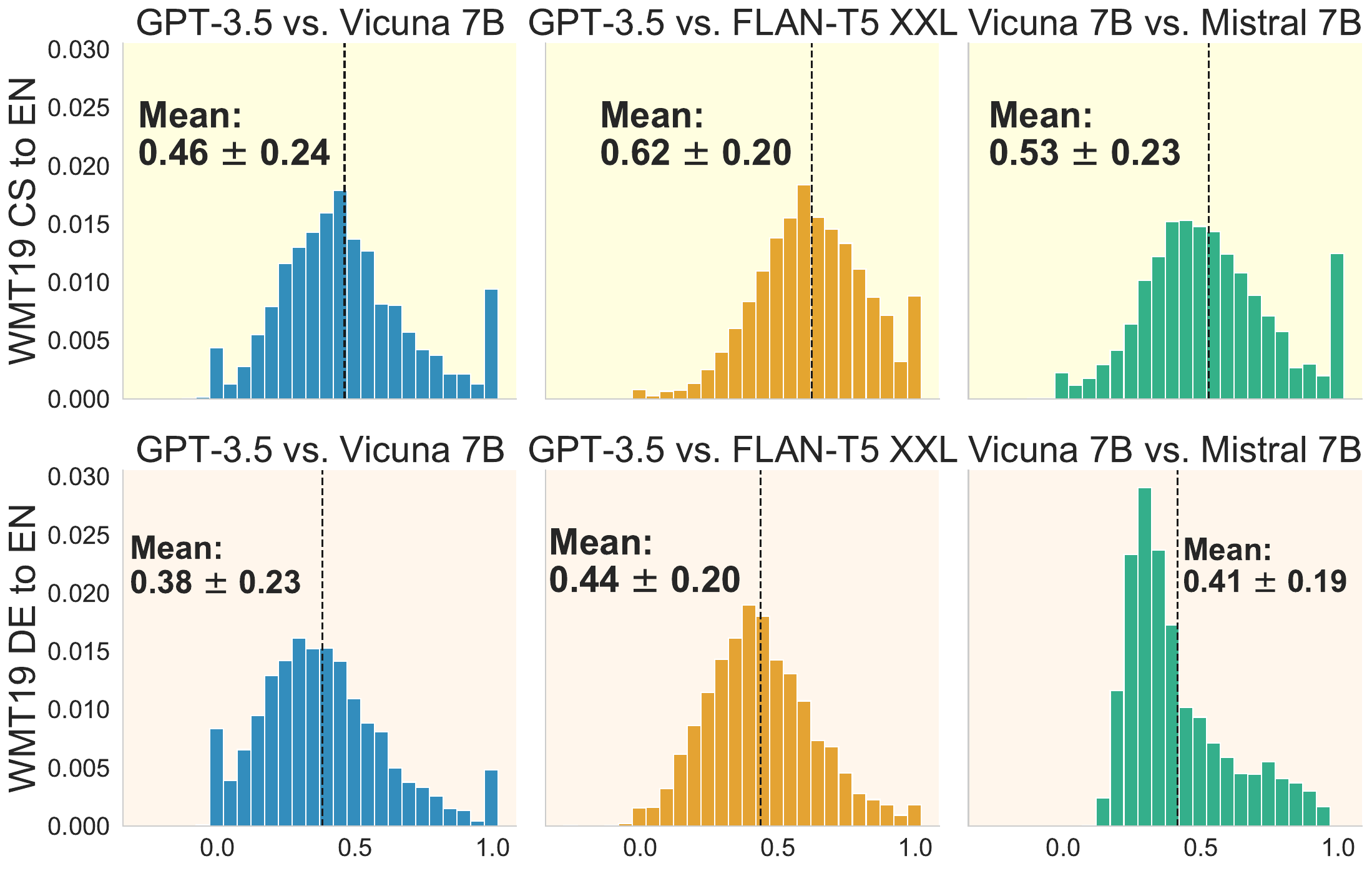}
     \end{subfigure}
        \caption{Histograms of \separability\ distributions for \textbf{summarization} (Left) and \textbf{translation} (Right). 
        For similar model pairs, CNN/DailyMail for news summarization and translation from a high-resource language (German) have lower average \separability\ compared to SAMSum for dialogue summarization and translation from a lower-resource language (Czech).
        We use length-adjusted BERTScore \cite{zhangbertscore} (defined in Section~\ref{sec:sim-fcns}) as the similarity metric for summarization and BLEU\cite{papineni-etal-2002-bleu} for translation.
        }
        \label{fig:sep-dists}
\end{figure*}

\begin{figure}[h]
\centering
    \includegraphics[width=\linewidth]{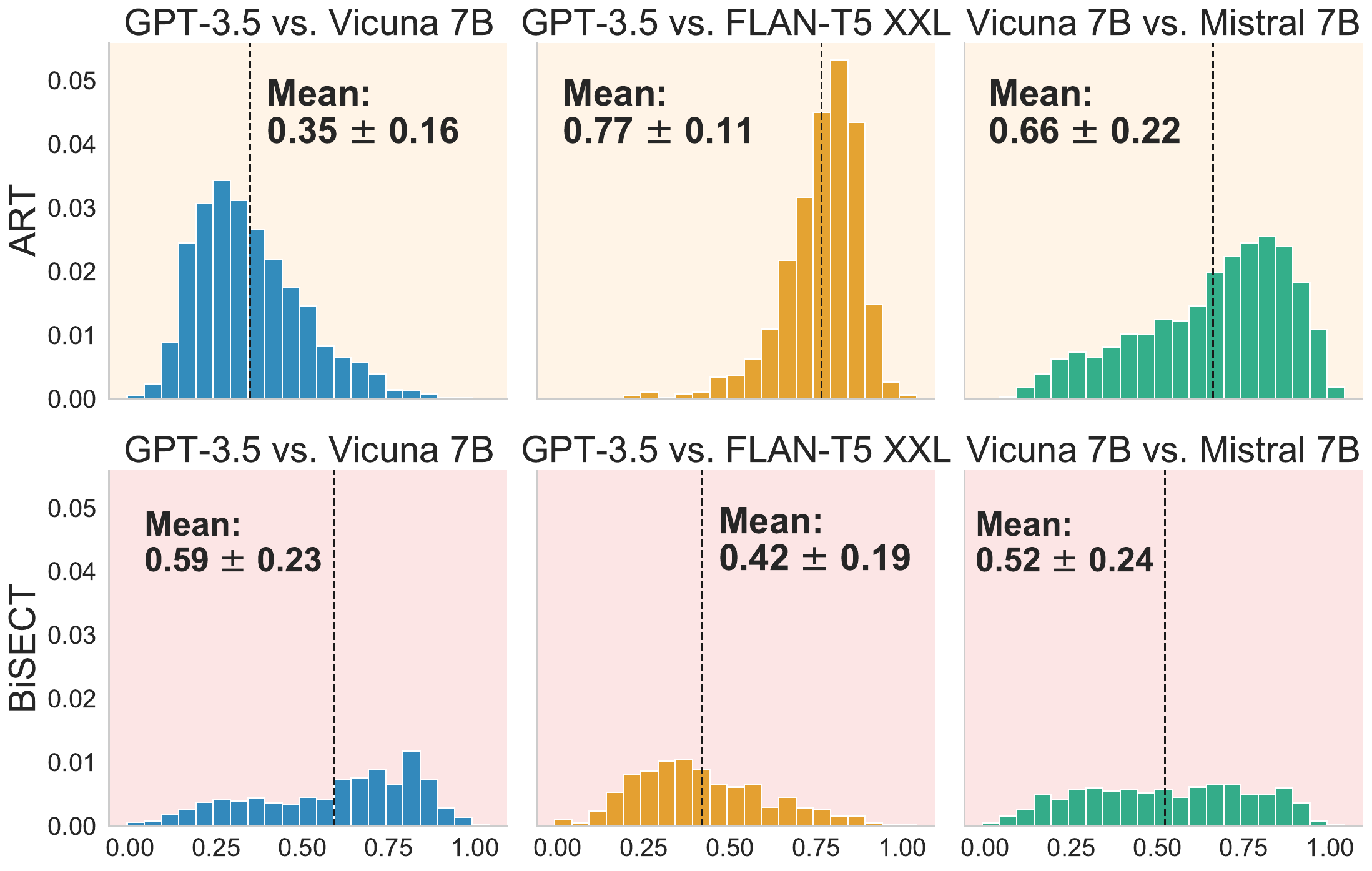}
    \vspace*{-6mm}
  \caption{
  \separability\ distributions for ART and BiSECT. We use length-adjusted BERTScore here (defined in Section~\ref{sec:sim-fcns}) as the similarity metric. \separability\ has higher variance, especially for BiSECT, largely caused by differences in instruction prompt interpretation; see \autoref{appendix:examples}.  
  \label{fig:other-sep-dist}
  }
\end{figure}

\subsection{Computing \separability on Generation Benchmarks}
\label{sec:estimating}

We compute \separability for various generation tasks under 6 different benchmarks using 3 model pairs, and demonstrate how \separability\ can allow model developers and users to visualize and understand how much a model pair's generations differ on a particular dataset. 
\autoref{fig:sep-dists} shows the empirical \separability\ distributions on two summarization benchmarks (left): CNN/DailyMail \citep{nallapati-etal-2016-abstractive} and SAMSum \citep{gliwa-etal-2019-samsum}, and two machine translation benchmarks (right):  Czech to English and German to English from the WMT-19 dataset \citep{barrault-etal-2019-findings}.
\autoref{fig:other-sep-dist} shows the empirical \separability\ distributions for abductive reasoning and sentence simplification, where we use the ART \citep{bhagavatula2019abductive} and BiSECT \citep{kim-etal-2021-bisect} benchmarks respectively.
\looseness=-1

For each benchmark, we compare three model pairs: \gpt\ vs. \vicuna\ \citep{vicuna_blog}, \vicuna\ vs. \mistral\ \citep{jiang2023mistral}, and \gpt\ vs. \flan\ \citep{longpre2023flan}.
We use identical instructions for each model, with different model-specific system prompts that were used to fine-tune each model during instruction tuning. 
We prompt each model in a zero-shot manner.
In \autoref{appendix:instructions}, we present the instruction prompts we used for each dataset described in this section.
For our experiments, we use temperature sampling with temperature $\tau = 0.5$.
To calculate alignment scores, we use $K=5$ samples and $C=25$ cross-alignment comparisons, unless mentioned otherwise. 
In \autoref{appendix:examples}, we present examples of low and high \separability\ generations corresponding to each of these tasks. 

\looseness=-1
We highlight several key takeaways.
Models with very different training methods, such as \gpt\ and \flan, output generations that are, on average, much easier to distinguish than models that are trained similarly, such as \vicuna\ and \mistral. 
Benchmarks such as CNN/DailyMail (Figure~\ref{fig:sep-dists}, top left) have instances with very low \separability\ on average (except \gpt\ versus \flan). 
These findings corroborate prior work that suggests CNN/DailyMail may not be useful for comparing modern LLMs \cite{goyal2022news, zhang2023benchmarking}.

Likewise, for machine translation, we see that it is easier to distinguish LLMs on lower-resource language test sets such as Czech$\rightarrow$English, compared to high-resource language test sets such as German$\rightarrow$English. \looseness=-1

Notably, \separability distributions for BiSECT are far less peaked (\autoref{fig:other-sep-dist}), indicating highly variable \separability. 
For both ART and BiSECT, differences in how the models interpreted the instructions (which didn't include explicit length constraints for these benchmarks) led to large differences in generation lengths, contributing to the high \separability of certain instances.
See \autoref{tab:qual-examples-high-other} in \autoref{appendix:examples} for examples.\looseness=-1

\begin{figure}[t!]
     \centering
\includegraphics[width=\linewidth]{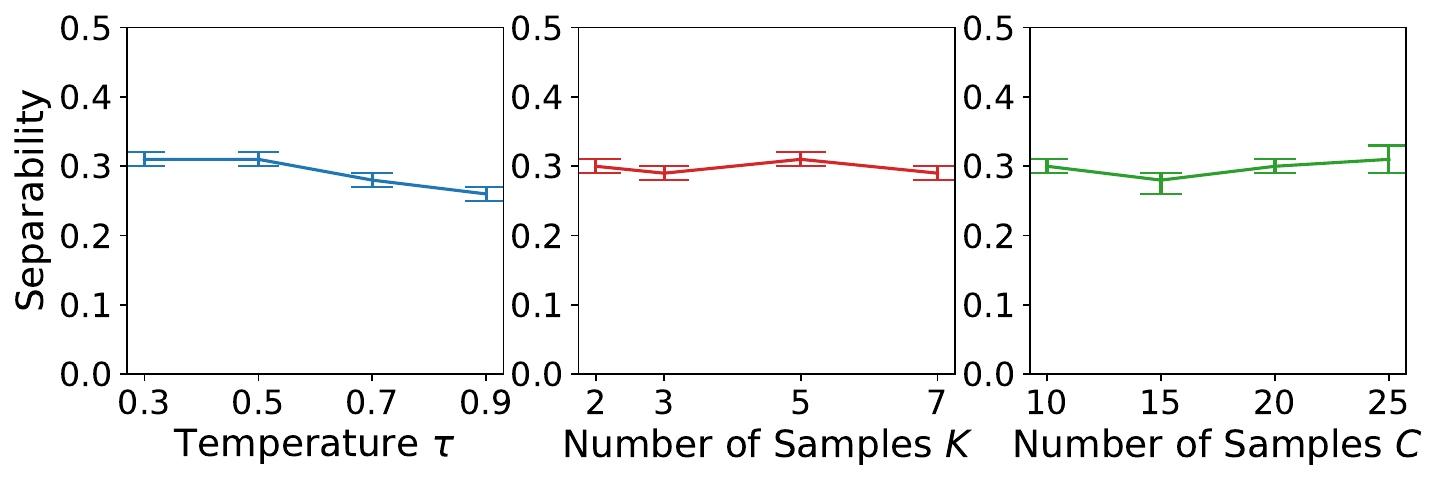}
\vspace*{-6mm}
\caption{\separability is robust to changes in the temperature $\tau$ used for generation (Left), the number of samples used to estimate alignments $K$ (Middle), and the number of cross-alignment comparisons $C$ (Right), for \gpt vs. \vicuna on SAMSum.
}
\label{fig:ablations}
\end{figure}
Our formulation of \separability is robust to the choice of  hyperparameters: $K$ for number of samples, $\tau$ for temperature in sampling and the number of samples used in computing cross-alignment, $C$; \autoref{fig:ablations} shows these ablations.

\section{\separability as Rating Consistency}
\label{sec:human-study}
We conduct a human study to verify our formulation for \separability as a meta-evaluation measure of preference rating consistency for a given instance $\mathbf{x}^i$ and a pair of generative models, $m_A$ and $m_B$.
Given that generation sets corresponding to low \separability\ instances are harder to distinguish, we hypothesize that preference judgments from raters on those sets will be inconsistent.
In other words, raters will not consistently prefer the same model's generation for any pair of generations sampled from low \separability instances.
On the other hand, we hypothesize that a rater's preference judgments on generation sets corresponding to high \separability instances will be consistent. 
While the inherent subjectivity present in generative evaluation may prevent raters from agreeing \emph{with each other} on high \separability instances, we hypothesize that each individual rater's judgments will consistently favor one model's generations. 

\begin{figure*}[h]
     \centering
     \includegraphics[width=\textwidth]{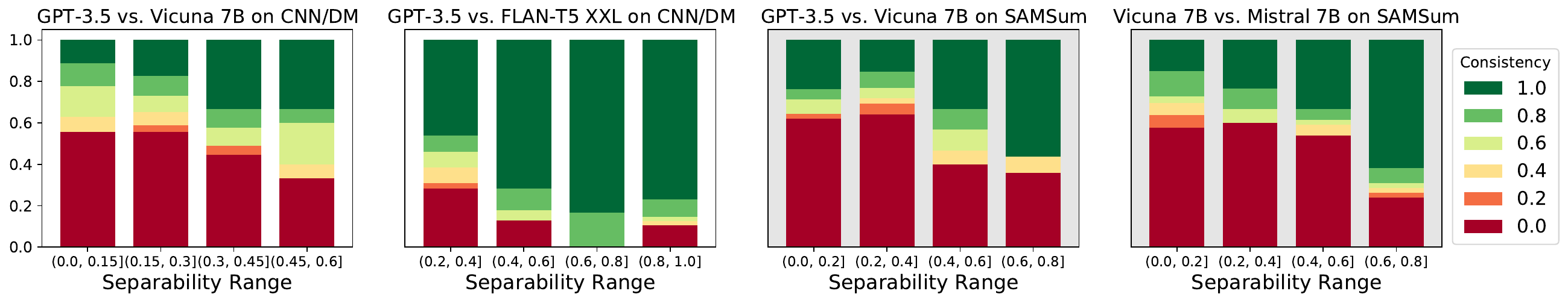}
     \vspace*{-6mm}
            \caption{Proportion of rating sets with each value in the range of consistency 
            corresponding to different \separability ranges. Ratings are not aggregated across raters for a test instance here. For each model pair and dataset configuration, the support is divided into equal sized ranges. The proportion of perfectly consistent ratings increases, and the proportion of inconsistent ratings decreases in higher \separability ranges.
            }
        \label{fig:con_ratings}
\end{figure*}

\subsection{Rating Consistency}
\label{sec:rating-con}
We define consistency of preference judgments as the average ratings from raters over $N$ sampled pairs.
For an input instance $\mathbf{x}^i$,
we sample $N$ generations $\mathbf{y}^{i, j} \in \mathcal{Y}, j \in 1\ldots N$ each from models $A$ and $B$ to obtain a set of paired generations $\mathcal{P}_{AB}(\mathbf{x}^i) = \left\{ \left(\mathbf{y}^{i,j}_A, \mathbf{y}^{i,j}_B\right)\right\}_{j=1}^N$.\looseness=-1

We represent a rater (annotator) $a$ by a rating function $r_a: \mathcal{Y} \times \mathcal{Y} \to \{-1, 0, 1\}$ that, for a pair of generations from $m_A$ and $m_B$, indicates which model's generation was preferred: $-1$ if $m_A$'s generation was preferred, $1$ in case $m_B$ was preferred, and $0$ if the rater had no preference. 
By having rater $a$ make preference judgments for each generation pair in $\mathcal{P}(\mathbf{x}^i)$, we obtain a rating set $\mathcal{R}_a(\mathbf{x}^i) := \left\{ r_a\left(\mathbf{y}^{i,j}_A, \mathbf{y}^{i,j}_B\right)\right\}_{j=1}^N$. 
We define the \textbf{\emph{consistency}}, $c\left(\mathcal{R}_a(\mathbf{x}^i)\right)$ of that rating set as: 
\begin{equation}
\begin{cases} 
    0, & \text{if } \{-1,1\} \subset \mathcal{R}_a(\mathbf{x}^i), \\
    \text{mean}( \left|r_a(\mathbf{x}^i)\right|_{r_a \in \mathcal{R}_a}), &\text{otherwise}
\end{cases}
\label{eq:consistency}
\end{equation}

Intuitively, if the rater prefers generations from both models during the course of the $N$ trials, we deem their rating set inconsistent (i.e. $c\left(\mathcal{R}_a(\mathbf{x}_i)\right)=0$). 
If the rater only ever picks one model's generations or $0$ ratings (ties), we deem their rating sets that include fewer $0$ ratings as more consistent.\looseness=-1 
 
In some cases, we may want to differentiate cases where there are differing degrees of inconsistency.
We address these cases through an additional metric called \textit{system preference strength}, with definitions and results in \autoref{appendix:strength-pref}.\looseness=-1

\subsection{Study Protocol and Settings}
\label{sec:study-protocol}
We conducted a human study with raters hired from Amazon Mechanical Turk.
Each human intelligence task (HIT) consisted of reading a source text (in our case, a news article or a dialogue) and $N=5$ pairs of generated summaries.\footnote{While we performed our experiments with summarization, we expect our results to hold for other tasks as well. In addition, the raters are performing $5$ cross-model comparisons as opposed to $25$ when calculating \separability, but we find that $5$ comparisons suffice.} 
For each summary pair, raters were asked to select which summary they preferred, with the option of picking no preference.

We hired a pool of 30 raters (workers) from Amazon Mechanical Turk, all of whom were native English speakers. 
Each rater was hired based on participation in a qualification study.
The raters were paid at a rate of \$1.20 per HIT, which was equal to roughly \$18 per hour using internally calculated time estimates for a single HIT.
The order in which models' summaries were shown in each pair was randomized in order to prevent positional bias. 
The HIT interface can be found in \autoref{appendix:hit-interface}.

Each HIT batch comprised source texts and summaries corresponding to a different model pair and dataset configuration.
We chose these configurations such that we had one set of instances with low average \separability ($\sim 0.2$), one with high average \separability ($\sim 0.7$), and two in-between:
\begin{compactenum}
    \item 
    {Low}: \gpt\ vs. \vicuna\ on CNN/DM 
    \item 
    {High}: \gpt\ vs. \flan\  on CNN/DM
    \item 
    {Medium}: \gpt\ vs. \vicuna\ on SAMSum
    \item 
    {Medium}: \vicuna\ vs. \mistral\ on SAMSum
\end{compactenum}
We collected ratings for 50 HITs for each of the four configurations. 
Since each HIT was rated by 3 raters, we have $600$ total rating sets. 

\begin{figure*}[ht]
     \centering
     \includegraphics[width=\textwidth]{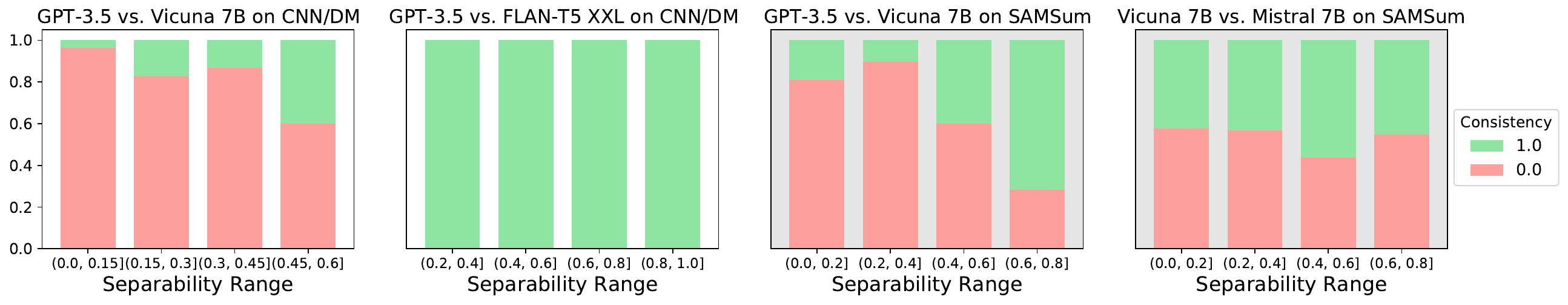}
     \vspace*{-6mm}
        \caption{Auto-raters from AlpacaEval produce more consistent ratings at higher \separability instances, much like human raters in Figure~\ref{fig:con_ratings}.
        }
        \label{fig:con_ratings_alpaca}
\end{figure*}

\begin{figure}[h]
\centering
    \includegraphics[width=0.9\linewidth]{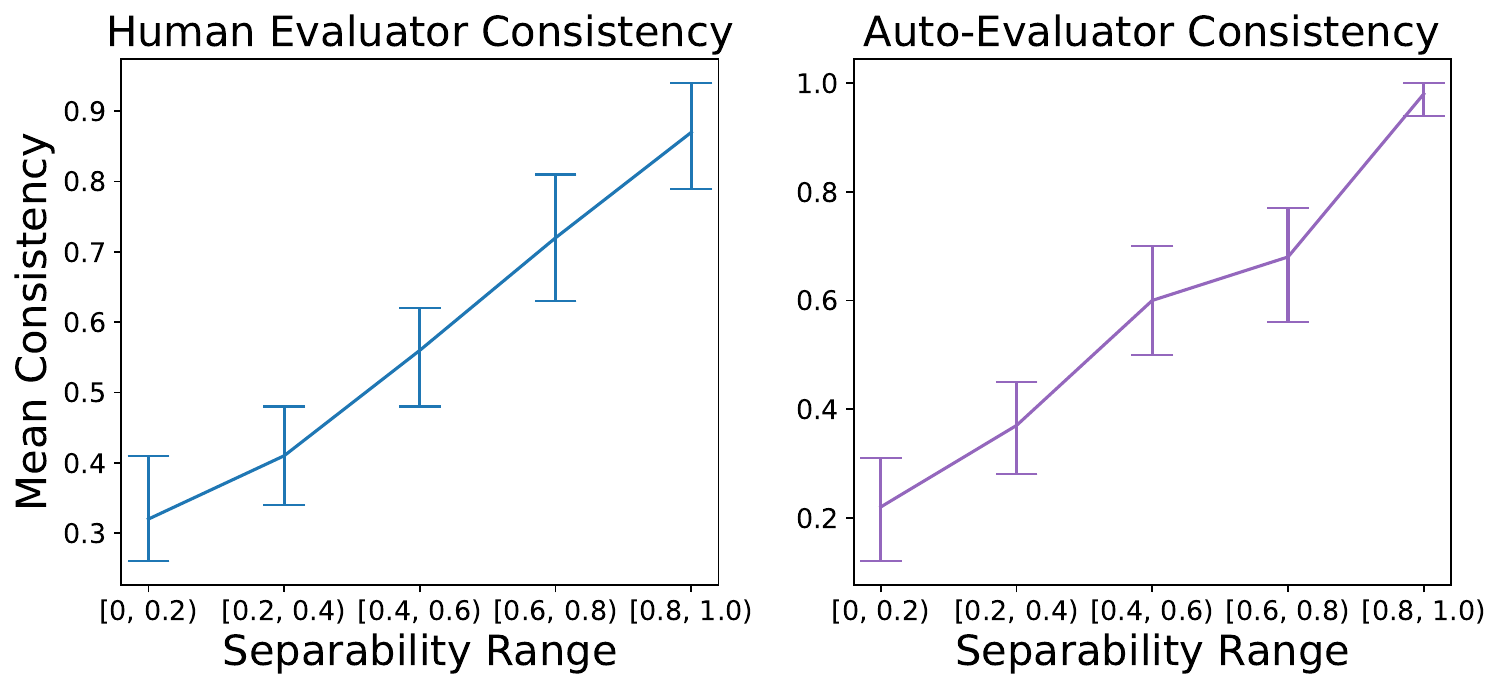}
    \vspace*{-2mm}
  \caption{Mean consistency of human and auto-rater preference judgments increases with \separability. Mean consistency is computed over all $600$ HITs collected. Consistency for a particular test instance is aggregated over raters by taking the mean of each individual rater's rating consistency.
  \label{fig:mean-consistency}
  }
\end{figure}

\subsection{Higher \separability\ Instances Receive Consistent Human Ratings}
\label{sec:human-eval-results}

To analyze the relationship between \separability ranges and consistency ratings, we bin the support of a \separability distribution for each our selected configurations into 
four equal-width bins. 
We plot the proportion of rating sets with each possible consistency value in each bin in \autoref{fig:con_ratings}.
For each model pair on the two benchmarks, we observe that, as \separability increases, the proportion of inconsistent rating sets decreases and the proportion of perfectly consistent ratings increases. 
For \separability $\delta_{A,B} \leq 0.2$, the majority of ratings are inconsistent. 
When $\delta_{A,B} \approx 0.4$, inconsistent ratings make up less than half of all ratings for all configurations.
Ratings for SAMSum tend to be more inconsistent across all ranges.  
\gpt and \flan, two models with different architectures and capabilities always produce more consistent ratings, even at lower \separability ranges. 
Nonetheless, there are a non-trivial number of inconsistent ratings at the lowest \separability range, and perfectly consistent ratings make up less than half of all ratings at this range. 
These findings indicate that \textbf{at higher values of \separability, raters are likely to give more reliable preference ratings} that are not dependent on the choice of generation pair that they are shown.

\subsection{Higher \separability Instances Receive Consistent Auto-Rater Ratings}
\label{sec:auto-eval-results}

LLM-based automatic raters have been rising in popularity \citep{chang2024survey} and are being used to replace human raters in many preference evaluation setups.
We ask: do auto-raters produce similar patterns of consistency as humans when making preference judgments?
We repeat our experiments using the same 600 instances in \S\ref{sec:human-eval-results} with auto-raters provided by AlpacaEval \citep{dubois2024length}
Each test instance is judged by three auto-raters, which have the highest correlations with humans (as of June 2024).\footnote{These models are: \texttt{alpaca\char`_eval\char`_gpt4}, \texttt{alpaca\char`_eval\char`_cot\char`_gpt4 \char`_turbo\char`_fn}, and \texttt{alpaca\char`_eval\char`_llama3\char`_70b\char`_fn} from AlpacaEval \citep{dubois2024length}.}
Since these raters cannot give tie judgments, the only possible consistencies are $0$ or $1$.

Results in Figure~\ref{fig:con_ratings_alpaca} show that, much like humans, auto-raters produce inconsistent ratings for low \separability instances under most configurations. 
For the \gpt\ vs. \flan\ comparison, the auto-raters always choose \gpt, whereas humans sometimes choose \flan\ in lower separability ranges. 
This phenomenon may be due to auto-raters being biased towards generations from their own model family \citep{panickssery2024llm}.\footnote{In our case, two of the auto-raters are in the \texttt{GPT} family.}
In contrast to human raters, auto-raters provide inconsistent ratings between \vicuna and \mistral even under higher \separability. 
This suggests that the factors influencing human judgments can be subtle and different from  those influencing auto-raters.

In Figure~\ref{fig:mean-consistency}, we plot the mean consistency for each of five equal-sized \separability ranges, aggregating over all four model pair and dataset configurations. 
Consistency increases with \separability for both human- and auto-raters, highlighting that raw \separability values can be directly compared across model pairs and datasets. 
Moreover, human- and auto-rater consistency patterns bear close resemblance with each other, with auto-rater consistency being slightly lower on average.
This resemblance suggests that \textbf{\separability is a valid meta-evaluation measure of the reliability of preference ratings, regardless of the type of rater}.\looseness=-1

\section{Applying \separability\ to ELO}
\label{sec:applying-sep}
As another concrete application of \separability, we investigate extending a popular novel method for ranking LLMs: \ELO ratings \citep{chiang2024chatbot, boubdir2023elo}.
In particular, we weight how much a new preference comparison affects a model's \ELO rating using the \separability of the test instance for that comparison. 

ELO ratings have emerged as a popular method of scoring and comparing LLMs \citep{chiang2024chatbot, boubdir2023elo}.
Originally developed to score and rank Chess players, \ELO ratings model the expected win probability of a model in a pairwise comparison.
After observing the outcome of a comparison between two models, both models' ratings are updated.
The \ELO updated rating for a model $m_A$ is given by
\begin{equation}
\centering
\ELO^{'}_A = \ELO_A + K^i(S_A^i - E_A^i),
\label{eq:elo-update}
\end{equation}
where $\ELO_A$ is the original rating, $S_A^i$ is the outcome of the comparison with instance $i$, $E_A^i$ is the expected win probability (based on the current \ELO score), and $K^i$ is a weighting factor which determines how much more recent comparisons should influence the rating.
The value of $S_A^i$ is equal to $1$ for a win, $0$ for a loss, and $\frac{1}{2}$ for a tie. 
Typically, $K^i$ is set to small values such as $K^i = 4$ for all $i$ in LLM comparisons \cite{chiang2024chatbot}; larger $K^i$ values are used in sports.

We propose incorporating \separability into the \ELO update in Equation~\ref{eq:elo-update} by modifying the weight $K^i$ for each new comparison based on its \separability value.
For an update $\ELO^{'}_A$ after a comparison on an instance $\mathbf{x}^i$ with \separability $\delta^i_{A, B}$, we use the weighting factor:
\begin{equation}
\centering
K_{\textsc{SEP}}^{i} = K^i \cdot \frac{\alpha}{1 + \exp\left(-\beta(\delta^i_{A, B}-T)\right)},
\label{eq:elo-update-sep}
\end{equation}
where $T$ is a chosen threshold, $\alpha$ and $\beta$ are hyperparameters controlling the how much the weight is updated and how fast.
We set $T = 0.4$ and $\alpha = 2$ and $\beta = 6$ in our experiments.\footnote{Since we do not have ground-truth regarding true model rankings, these parameters are dependent on user preference}
Intuitively, this update rule upweights $K^i$ (Equation~\ref{eq:elo-update}) when the input $i$ has high \separability, and vice-versa.
When the input's \separability value is at the chosen threshold $T$, $K$ is not updated.

\begin{figure}[t]
\centering
    \includegraphics[width=0.9\linewidth]{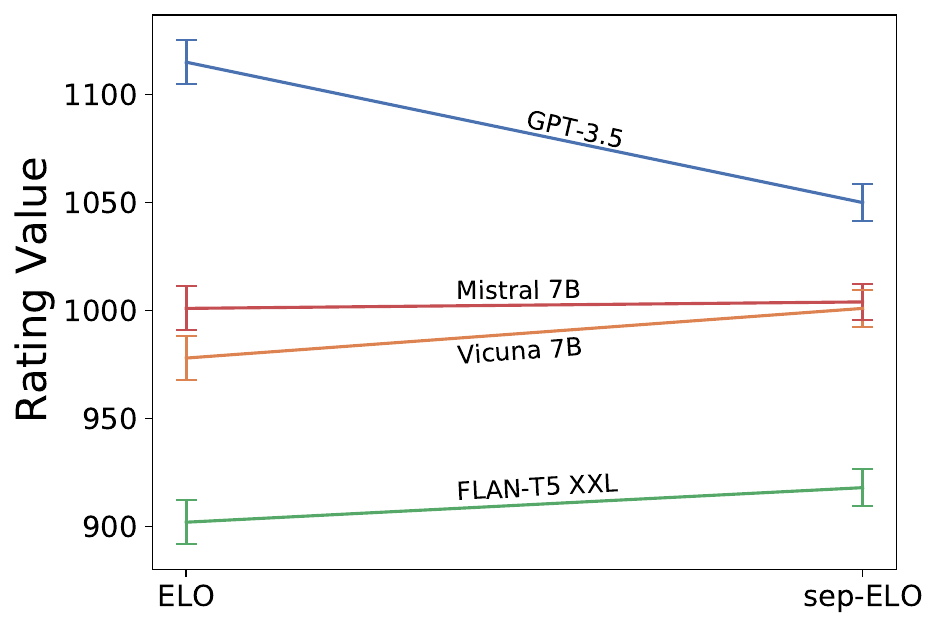}
  \vspace*{-4mm}
  \caption{After incorporating \separability into \ELO, we get narrower gaps in model rankings, reflecting similar capabilities of both \mistral and \vicuna.
  \label{fig:sep-elo}
  }
\end{figure}

We compute \ELO and \separability-weighted \ELO (\sepELO; Equation~\ref{eq:elo-update-sep}) using data from our 600 human evaluation HITs as preference judgments (\S\ref{sec:human-study}). 
To calculate these ratings, we sample one rating for each input from our pool of ratings.
We compute confidence intervals with the bootstrap method with $100$ trials.
\autoref{fig:sep-elo} shows that \sepELO has narrower gaps in model ranking, suggesting that models are more similar under adjustments to consistency of judgments (or, \separability).
We acknowledge that our results are a pilot due to the limited number of ratings we could use in our computation.
However, we expect \sepELO can reveal reliable trends even when applied to larger sets of preference data such as LMSYS\footnote{\url{https://lmsys.org/}}. 
Further testing with larger sets of ratings can also help users more optimally tune the hyperparameters used in the \separability-ELO score calculation. 
However, we note that hyperparameter selection involves a subjective judgment on how much a practitioner wishes to incorporate \separability values into the score update. 

\paragraph{Alternative Applications of \separability}
Beyond \sepELO, \separability values could be used for adversarially filtering test sets \cite{bras2020adversarial}.
Not only would this lead to fine-grained comparisons between models, but could also lead to obtaining cost- and time-efficient human ratings.
However, some caution is to be urged since such filtering may lead to biases \cite{schwartz-stanovsky-2022-limitations}, since low \separability instances can still contain valuable information. 
Instead, we recommend importance \emph{weighting} instances by \separability when sampling instances for human judgments, in a similar manner as it is used in \ELO ratings. 
That is, we recommend that high \separability instances be used for human preference judgment collection.
Alternatively, a stratified sampling approach from different separability ranges could ensure a more robust preference evaluation scheme.

\section{Related Work}
\label{sec:related-work}

\paragraph{Model Output Variability}
\label{sec:variability}
\citet{giulianelli-etal-2023-comes} comprehensively characterize LLM vs human output variability, with a focus on comparing it to human output variability.
\citet{suzgun2022follow,bertsch-etal-2023-mbr} take advantage of production variability to select more optimal generations using Minimum Bayes Risk (MBR) decoding.
In a similar vein, our work incorporates variability in generations into our meta-evaluation measure.

\paragraph{Prioritizing Test Instances}
\label{sec:prioritizing}
\citet{rodriguez-etal-2021-evaluation, vania-etal-2021-comparing} evaluate test instances on a variety of dimensions such as difficulty and discriminability (similar to our notion of \separability) using Item Response Theory (IRT), albeit in a text classification setting.
\citet{boubdir2023prompts, ashury2024label} study prioritizing test instances for human evaluation for efficiency-related purposes. 
However, the approach of \citet{boubdir2023prompts} relies on access to model logits which are not necessarily available to LLM users. 
Moreover, we take a more task-centered approach. 
\citet{ashury2024label} use embeddings of model outputs and a clustering-based method to select a subset of test instances that are most illustrative of model differences.

\section{Conclusion}
\label{sec:conclusion}
We present \separability, a meta-evaluation measure that estimates how suitable a test instance is for pairwise preference elicitation. 
We show that instances with high \separability yield more consistent human judgments. 
We show that the test distribution of \separability can be used to analyze how useful a benchmark may be for the comparison of two LLMs.  
We show that \separability can be incorporated into \ELO scores.
Our work shows that \separability can help LLM developers and users determine and prioritize evaluation instances and benchmarks.
Future work will look at applying \separability in building quality filters for preference tuning data for learning from human feedback.\looseness=-1

\clearpage

\section*{Acknowledgments}
We would like to thank several members of USC NLP for their valuable feedback, insights, and help in designing and testing our human evaluation: Jaspreet Ranjit, Risha Surana, Xinyue Cui, Yoonsoo Nam, Keyu He, Joseph Liu, Matthew Finlayson, Brihi Joshi, Johnny Wei, and Robin Jia.
This research was supported in part by a Young Investigator award from the Allen Institute for AI, as well as the National Science Foundation under Grant No. IIS-2403436.
Any opinions, findings, and conclusions or recommendations expressed in this material are those of the author(s) and do not necessarily reflect the views of the National Science Foundation.

\section*{Limitations}
\label{sec:Limitations}
%\swabha{only so many human comparisons, only on summarization. ELO experiments were also limited by availability of data. only english. only 5 human comparisons per pair, only 5 pairs. while we don't propose autoeval, we urge caution against auto eval which may introduce biases.}

We only used \separability in tasks that produce English output generations.
Due to resource and time constraints, our human evaluation for verifying \separability was done on two summarization tasks with five summary pair comparisons for each instance by three annotators. 
We chose instances with separability values for our human comparisons to highlight different levels of consistency in ratings.
We expect that an even larger comparison would reveal more fine-grained variations.
Our analysis on applying to \separability to ELO also used limited human comparisons and model pairs.
Larger scale preference data collection would be needed for more fine-grained analysis. 
While we expect our conclusions to hold for different tasks, different similarity functions may be optimal for different tasks, since the importance of \emph{what} type of differences are most influential for human judgments can vary by task.
Furthermore, we only used $5$ human comparisons per pair and $5$ samples to compute \separability for our main experiments.

\bibliography{anthology,citations}

\begin{thebibliography}{39}
\expandafter\ifx\csname natexlab\endcsname\relax\def\natexlab#1{#1}\fi

\bibitem[{Amidei et~al.(2019)Amidei, Piwek, and Willis}]{amidei-etal-2019-agreement}
Jacopo Amidei, Paul Piwek, and Alistair Willis. 2019.
\newblock \href {https://doi.org/10.18653/v1/W19-8642} {Agreement is overrated: A plea for correlation to assess human evaluation reliability}.
\newblock In \emph{Proceedings of the 12th International Conference on Natural Language Generation}, pages 344--354, Tokyo, Japan. Association for Computational Linguistics.

\bibitem[{Ashury-Tahan et~al.(2024)Ashury-Tahan, Sznajder, Choshen, Ein-Dor, Shnarch, and Gera}]{ashury2024label}
Shir Ashury-Tahan, Benjamin Sznajder, Leshem Choshen, Liat Ein-Dor, Eyal Shnarch, and Ariel Gera. 2024.
\newblock \href {https://arxiv.org/abs/2402.07891} {Label-efficient model selection for text generation}.
\newblock \emph{arXiv preprint arXiv:2402.07891}.

\bibitem[{Barrault et~al.(2019)Barrault, Bojar, Costa-juss{\`a}, Federmann, Fishel, Graham, Haddow, Huck, Koehn, Malmasi, Monz, M{\"u}ller, Pal, Post, and Zampieri}]{barrault-etal-2019-findings}
Lo{\"\i}c Barrault, Ond{\v{r}}ej Bojar, Marta~R. Costa-juss{\`a}, Christian Federmann, Mark Fishel, Yvette Graham, Barry Haddow, Matthias Huck, Philipp Koehn, Shervin Malmasi, Christof Monz, Mathias M{\"u}ller, Santanu Pal, Matt Post, and Marcos Zampieri. 2019.
\newblock \href {https://doi.org/10.18653/v1/W19-5301} {Findings of the 2019 conference on machine translation ({WMT}19)}.
\newblock In \emph{Proceedings of the Fourth Conference on Machine Translation (Volume 2: Shared Task Papers, Day 1)}, pages 1--61, Florence, Italy. Association for Computational Linguistics.

\bibitem[{Bertsch et~al.(2023)Bertsch, Xie, Neubig, and Gormley}]{bertsch-etal-2023-mbr}
Amanda Bertsch, Alex Xie, Graham Neubig, and Matthew Gormley. 2023.
\newblock \href {https://doi.org/10.18653/v1/2023.bigpicture-1.9} {It{'}s {MBR} all the way down: Modern generation techniques through the lens of minimum {B}ayes risk}.
\newblock In \emph{Proceedings of the Big Picture Workshop}, pages 108--122, Singapore. Association for Computational Linguistics.

\bibitem[{Bhagavatula et~al.(2019)Bhagavatula, Le~Bras, Malaviya, Sakaguchi, Holtzman, Rashkin, Downey, Yih, and Choi}]{bhagavatula2019abductive}
Chandra Bhagavatula, Ronan Le~Bras, Chaitanya Malaviya, Keisuke Sakaguchi, Ari Holtzman, Hannah Rashkin, Doug Downey, Wen-tau Yih, and Yejin Choi. 2019.
\newblock \href {https://openreview.net/forum?id=Byg1v1HKDB} {Abductive commonsense reasoning}.
\newblock In \emph{International Conference on Learning Representations}.

\bibitem[{Boubdir et~al.(2023{\natexlab{a}})Boubdir, Kim, Ermis, Fadaee, and Hooker}]{boubdir2023prompts}
Meriem Boubdir, Edward Kim, Beyza Ermis, Marzieh Fadaee, and Sara Hooker. 2023{\natexlab{a}}.
\newblock \href {https://arxiv.org/abs/2310.14424} {Which prompts make the difference? data prioritization for efficient human llm evaluation}.
\newblock \emph{arXiv preprint arXiv:2310.14424}.

\bibitem[{Boubdir et~al.(2023{\natexlab{b}})Boubdir, Kim, Ermis, Hooker, and Fadaee}]{boubdir2023elo}
Meriem Boubdir, Edward Kim, Beyza Ermis, Sara Hooker, and Marzieh Fadaee. 2023{\natexlab{b}}.
\newblock \href {https://arxiv.org/abs/2311.17295} {Elo uncovered: Robustness and best practices in language model evaluation}.
\newblock \emph{arXiv preprint arXiv:2311.17295}.

\bibitem[{Bras et~al.(2020)Bras, Swayamdipta, Bhagavatula, Zellers, Peters, Sabharwal, and Choi}]{bras2020adversarial}
Ronan~Le Bras, Swabha Swayamdipta, Chandra Bhagavatula, Rowan Zellers, Matthew Peters, Ashish Sabharwal, and Yejin Choi. 2020.
\newblock \href {https://openreview.net/forum?id=H1g8p1BYvS} {Adversarial filters of dataset biases}.
\newblock In \emph{Proc. of ICML}.

\bibitem[{Chang et~al.(2024)Chang, Wang, Wang, Wu, Yang, Zhu, Chen, Yi, Wang, Wang, Ye, Zhang, Chang, Yu, Yang, and Xie}]{chang2024survey}
Yupeng Chang, Xu~Wang, Jindong Wang, Yuan Wu, Linyi Yang, Kaijie Zhu, Hao Chen, Xiaoyuan Yi, Cunxiang Wang, Yidong Wang, Wei Ye, Yue Zhang, Yi~Chang, Philip~S. Yu, Qiang Yang, and Xing Xie. 2024.
\newblock \href {https://doi.org/10.1145/3641289} {A survey on evaluation of large language models}.
\newblock \emph{ACM Trans. Intell. Syst. Technol.}, 15(3).

\bibitem[{Chiang et~al.(2024)Chiang, Zheng, Sheng, Angelopoulos, Li, Li, Zhang, Zhu, Jordan, Gonzalez et~al.}]{chiang2024chatbot}
Wei-Lin Chiang, Lianmin Zheng, Ying Sheng, Anastasios~Nikolas Angelopoulos, Tianle Li, Dacheng Li, Hao Zhang, Banghua Zhu, Michael Jordan, Joseph~E Gonzalez, et~al. 2024.
\newblock \href {https://arxiv.org/abs/2403.04132} {Chatbot arena: An open platform for evaluating llms by human preference}.
\newblock \emph{arXiv preprint arXiv:2403.04132}.

\bibitem[{Dubois et~al.(2024)Dubois, Galambosi, Liang, and Hashimoto}]{dubois2024length}
Yann Dubois, Bal{\'a}zs Galambosi, Percy Liang, and Tatsunori~B Hashimoto. 2024.
\newblock \href {https://arxiv.org/abs/2404.04475} {Length-controlled alpacaeval: A simple way to debias automatic evaluators}.
\newblock \emph{arXiv preprint arXiv:2404.04475}.

\bibitem[{Ethayarajh and Jurafsky(2022)}]{ethayarajh2022authenticity}
Kawin Ethayarajh and Dan Jurafsky. 2022.
\newblock \href {https://aclanthology.org/2022.emnlp-main.406} {The authenticity gap in human evaluation}.
\newblock In \emph{Proceedings of the 2022 Conference on Empirical Methods in Natural Language Processing}, pages 6056--6070, Abu Dhabi, United Arab Emirates. Association for Computational Linguistics.

\bibitem[{Gehrmann et~al.(2022)Gehrmann, Clark, and Sellam}]{gehrmann2022repairing}
Sebastian Gehrmann, Elizabeth Clark, and Thibault Sellam. 2022.
\newblock \href {https://arxiv.org/abs/2202.06935} {Repairing the cracked foundation: A survey of obstacles in evaluation practices for generated text}.
\newblock \emph{arXiv preprint arXiv:2202.06935}.

\bibitem[{Giulianelli et~al.(2023)Giulianelli, Baan, Aziz, Fern{\'a}ndez, and Plank}]{giulianelli-etal-2023-comes}
Mario Giulianelli, Joris Baan, Wilker Aziz, Raquel Fern{\'a}ndez, and Barbara Plank. 2023.
\newblock \href {https://doi.org/10.18653/v1/2023.emnlp-main.887} {What comes next? evaluating uncertainty in neural text generators against human production variability}.
\newblock In \emph{Proceedings of the 2023 Conference on Empirical Methods in Natural Language Processing}, pages 14349--14371, Singapore. Association for Computational Linguistics.

\bibitem[{Gliwa et~al.(2019)Gliwa, Mochol, Biesek, and Wawer}]{gliwa-etal-2019-samsum}
Bogdan Gliwa, Iwona Mochol, Maciej Biesek, and Aleksander Wawer. 2019.
\newblock \href {https://doi.org/10.18653/v1/D19-5409} {{SAMS}um corpus: A human-annotated dialogue dataset for abstractive summarization}.
\newblock In \emph{Proceedings of the 2nd Workshop on New Frontiers in Summarization}, pages 70--79, Hong Kong, China. Association for Computational Linguistics.

\bibitem[{Goyal et~al.(2022)Goyal, Li, and Durrett}]{goyal2022news}
Tanya Goyal, Junyi~Jessy Li, and Greg Durrett. 2022.
\newblock \href {https://arxiv.org/abs/2209.12356} {News summarization and evaluation in the era of gpt-3}.
\newblock \emph{arXiv preprint arXiv:2209.12356}.

\bibitem[{Jiang et~al.(2023)Jiang, Sablayrolles, Mensch, Bamford, Chaplot, Casas, Bressand, Lengyel, Lample, Saulnier et~al.}]{jiang2023mistral}
Albert~Q Jiang, Alexandre Sablayrolles, Arthur Mensch, Chris Bamford, Devendra~Singh Chaplot, Diego de~las Casas, Florian Bressand, Gianna Lengyel, Guillaume Lample, Lucile Saulnier, et~al. 2023.
\newblock \href {https://arxiv.org/abs/2310.06825} {Mistral 7b}.
\newblock \emph{arXiv preprint arXiv:2310.06825}.

\bibitem[{Kim et~al.(2021)Kim, Maddela, Kriz, Xu, and Callison-Burch}]{kim-etal-2021-bisect}
Joongwon Kim, Mounica Maddela, Reno Kriz, Wei Xu, and Chris Callison-Burch. 2021.
\newblock \href {https://doi.org/10.18653/v1/2021.emnlp-main.500} {{B}i{SECT}: Learning to split and rephrase sentences with bitexts}.
\newblock In \emph{Proceedings of the 2021 Conference on Empirical Methods in Natural Language Processing}, pages 6193--6209, Online and Punta Cana, Dominican Republic. Association for Computational Linguistics.

\bibitem[{Lin et~al.(2024)Lin, Deng, Chandu, Brahman, Ravichander, Pyatkin, Dziri, Bras, and Choi}]{lin2024wildbench}
Bill~Yuchen Lin, Yuntian Deng, Khyathi Chandu, Faeze Brahman, Abhilasha Ravichander, Valentina Pyatkin, Nouha Dziri, Ronan~Le Bras, and Yejin Choi. 2024.
\newblock \href {https://arxiv.org/abs/2406.04770} {Wildbench: Benchmarking llms with challenging tasks from real users in the wild}.
\newblock \emph{arXiv preprint arXiv:2406.04770}.

\bibitem[{Lin(2004)}]{lin-2004-rouge}
Chin-Yew Lin. 2004.
\newblock \href {https://aclanthology.org/W04-1013} {{ROUGE}: A package for automatic evaluation of summaries}.
\newblock In \emph{Text Summarization Branches Out}, pages 74--81, Barcelona, Spain. Association for Computational Linguistics.

\bibitem[{Liu et~al.(2023)Liu, Iter, Xu, Wang, Xu, and Zhu}]{liu-etal-2023-g}
Yang Liu, Dan Iter, Yichong Xu, Shuohang Wang, Ruochen Xu, and Chenguang Zhu. 2023.
\newblock \href {https://doi.org/10.18653/v1/2023.emnlp-main.153} {{G}-eval: {NLG} evaluation using gpt-4 with better human alignment}.
\newblock In \emph{Proceedings of the 2023 Conference on Empirical Methods in Natural Language Processing}, pages 2511--2522, Singapore. Association for Computational Linguistics.

\bibitem[{LMSys(2023)}]{vicuna_blog}
LMSys. 2023.
\newblock \href {https://lmsys.org/blog/2023-03-30-vicuna/} {Vicuna: A cloud-native computing service for machine learning workflows}.

\bibitem[{Longpre et~al.(2023)Longpre, Hou, Vu, Webson, Chung, Tay, Zhou, Le, Zoph, Wei et~al.}]{longpre2023flan}
Shayne Longpre, Le~Hou, Tu~Vu, Albert Webson, Hyung~Won Chung, Yi~Tay, Denny Zhou, Quoc~V Le, Barret Zoph, Jason Wei, et~al. 2023.
\newblock \href {https://proceedings.mlr.press/v202/longpre23a.html} {The flan collection: Designing data and methods for effective instruction tuning}.
\newblock In \emph{International Conference on Machine Learning}, pages 22631--22648. PMLR.

\bibitem[{Nallapati et~al.(2016)Nallapati, Zhou, dos Santos, Gul{\c{c}}ehre, and Xiang}]{nallapati-etal-2016-abstractive}
Ramesh Nallapati, Bowen Zhou, Cicero dos Santos, {\c{C}}a{\u{g}}lar Gul{\c{c}}ehre, and Bing Xiang. 2016.
\newblock \href {https://doi.org/10.18653/v1/K16-1028} {Abstractive text summarization using sequence-to-sequence {RNN}s and beyond}.
\newblock In \emph{Proceedings of the 20th {SIGNLL} Conference on Computational Natural Language Learning}, pages 280--290, Berlin, Germany. Association for Computational Linguistics.

\bibitem[{Panickssery et~al.(2024)Panickssery, Bowman, and Feng}]{panickssery2024llm}
Arjun Panickssery, Samuel~R Bowman, and Shi Feng. 2024.
\newblock \href {https://arxiv.org/abs/2404.13076} {Llm evaluators recognize and favor their own generations}.
\newblock \emph{arXiv preprint arXiv:2404.13076}.

\bibitem[{Papineni et~al.(2002)Papineni, Roukos, Ward, and Zhu}]{papineni-etal-2002-bleu}
Kishore Papineni, Salim Roukos, Todd Ward, and Wei-Jing Zhu. 2002.
\newblock \href {https://doi.org/10.3115/1073083.1073135} {{B}leu: a method for automatic evaluation of machine translation}.
\newblock In \emph{Proceedings of the 40th Annual Meeting of the Association for Computational Linguistics}, pages 311--318, Philadelphia, Pennsylvania, USA. Association for Computational Linguistics.

\bibitem[{Prabhakaran et~al.(2021)Prabhakaran, Mostafazadeh~Davani, and Diaz}]{prabhakaran-etal-2021-releasing}
Vinodkumar Prabhakaran, Aida Mostafazadeh~Davani, and Mark Diaz. 2021.
\newblock \href {https://doi.org/10.18653/v1/2021.law-1.14} {On releasing annotator-level labels and information in datasets}.
\newblock In \emph{Proceedings of the Joint 15th Linguistic Annotation Workshop (LAW) and 3rd Designing Meaning Representations (DMR) Workshop}, pages 133--138, Punta Cana, Dominican Republic. Association for Computational Linguistics.

\bibitem[{Rodriguez et~al.(2021)Rodriguez, Barrow, Hoyle, Lalor, Jia, and Boyd-Graber}]{rodriguez-etal-2021-evaluation}
Pedro Rodriguez, Joe Barrow, Alexander~Miserlis Hoyle, John~P. Lalor, Robin Jia, and Jordan Boyd-Graber. 2021.
\newblock \href {https://doi.org/10.18653/v1/2021.acl-long.346} {Evaluation examples are not equally informative: How should that change {NLP} leaderboards?}
\newblock In \emph{Proceedings of the 59th Annual Meeting of the Association for Computational Linguistics and the 11th International Joint Conference on Natural Language Processing (Volume 1: Long Papers)}, pages 4486--4503, Online. Association for Computational Linguistics.

\bibitem[{Schwartz and Stanovsky(2022)}]{schwartz-stanovsky-2022-limitations}
Roy Schwartz and Gabriel Stanovsky. 2022.
\newblock \href {https://doi.org/10.18653/v1/2022.findings-naacl.168} {On the limitations of dataset balancing: The lost battle against spurious correlations}.
\newblock In \emph{Findings of the Association for Computational Linguistics: NAACL 2022}, pages 2182--2194, Seattle, United States. Association for Computational Linguistics.

\bibitem[{Sun et~al.(2019)Sun, Shapira, Dagan, and Nenkova}]{sun-etal-2019-compare}
Simeng Sun, Ori Shapira, Ido Dagan, and Ani Nenkova. 2019.
\newblock \href {https://doi.org/10.18653/v1/W19-2303} {How to compare summarizers without target length? pitfalls, solutions and re-examination of the neural summarization literature}.
\newblock In \emph{Proceedings of the Workshop on Methods for Optimizing and Evaluating Neural Language Generation}, pages 21--29, Minneapolis, Minnesota. Association for Computational Linguistics.

\bibitem[{Suzgun et~al.(2022)Suzgun, Melas-Kyriazi, and Jurafsky}]{suzgun2022follow}
Mirac Suzgun, Luke Melas-Kyriazi, and Dan Jurafsky. 2022.
\newblock \href {https://arxiv.org/pdf/2211.07634.pdf} {Follow the wisdom of the crowd: Effective text generation via minimum bayes risk decoding}.
\newblock \emph{arXiv preprint arXiv:2211.07634}.

\bibitem[{Tsvilodub et~al.(2024)Tsvilodub, Wang, Grosch, and Franke}]{tsvilodub2024predictions}
Polina Tsvilodub, Hening Wang, Sharon Grosch, and Michael Franke. 2024.
\newblock \href {https://arxiv.org/abs/2403.00998} {Predictions from language models for multiple-choice tasks are not robust under variation of scoring methods}.
\newblock \emph{arXiv preprint arXiv:2403.00998}.

\bibitem[{Vania et~al.(2021)Vania, Htut, Huang, Mungra, Pang, Phang, Liu, Cho, and Bowman}]{vania-etal-2021-comparing}
Clara Vania, Phu~Mon Htut, William Huang, Dhara Mungra, Richard~Yuanzhe Pang, Jason Phang, Haokun Liu, Kyunghyun Cho, and Samuel~R. Bowman. 2021.
\newblock \href {https://doi.org/10.18653/v1/2021.acl-long.92} {Comparing test sets with item response theory}.
\newblock In \emph{Proceedings of the 59th Annual Meeting of the Association for Computational Linguistics and the 11th International Joint Conference on Natural Language Processing (Volume 1: Long Papers)}, pages 1141--1158, Online. Association for Computational Linguistics.

\bibitem[{Wang et~al.(2023)Wang, Li, Chen, Cai, Zhu, Lin, Cao, Liu, Liu, and Sui}]{wang2023large}
Peiyi Wang, Lei Li, Liang Chen, Zefan Cai, Dawei Zhu, Binghuai Lin, Yunbo Cao, Qi~Liu, Tianyu Liu, and Zhifang Sui. 2023.
\newblock \href {https://arxiv.org/abs/2305.17926} {Large language models are not fair evaluators}.
\newblock \emph{arXiv preprint arXiv:2305.17926}.

\bibitem[{Wu and Aji(2023)}]{wu2023style}
Minghao Wu and Alham~Fikri Aji. 2023.
\newblock \href {https://arxiv.org/abs/2307.03025} {Style over substance: Evaluation biases for large language models}.
\newblock \emph{arXiv preprint arXiv:2307.03025}.

\bibitem[{Zeng et~al.(2023)Zeng, Yu, Gao, Meng, Goyal, and Chen}]{zeng2023evaluating}
Zhiyuan Zeng, Jiatong Yu, Tianyu Gao, Yu~Meng, Tanya Goyal, and Danqi Chen. 2023.
\newblock \href {https://openreview.net/forum?id=tr0KidwPLc} {Evaluating large language models at evaluating instruction following}.
\newblock In \emph{The Twelfth International Conference on Learning Representations}.

\bibitem[{Zhang et~al.(2019)Zhang, Kishore, Wu, Weinberger, and Artzi}]{zhangbertscore}
Tianyi Zhang, Varsha Kishore, Felix Wu, Kilian~Q Weinberger, and Yoav Artzi. 2019.
\newblock \href {https://arxiv.org/abs/1904.09675} {Bertscore: Evaluating text generation with bert}.
\newblock In \emph{International Conference on Learning Representations}.

\bibitem[{Zhang et~al.(2023)Zhang, Ladhak, Durmus, Liang, McKeown, and Hashimoto}]{zhang2023benchmarking}
Tianyi Zhang, Faisal Ladhak, Esin Durmus, Percy Liang, Kathleen McKeown, and Tatsunori~B Hashimoto. 2023.
\newblock \href {https://arxiv.org/abs/2301.13848} {Benchmarking large language models for news summarization}.
\newblock \emph{arXiv preprint arXiv:2301.13848}.

\bibitem[{Zheng et~al.(2024)Zheng, Chiang, Sheng, Zhuang, Wu, Zhuang, Lin, Li, Li, Xing et~al.}]{zheng2024judging}
Lianmin Zheng, Wei-Lin Chiang, Ying Sheng, Siyuan Zhuang, Zhanghao Wu, Yonghao Zhuang, Zi~Lin, Zhuohan Li, Dacheng Li, Eric Xing, et~al. 2024.
\newblock \href {https://proceedings.neurips.cc/paper_files/paper/2023/file/91f18a1287b398d378ef22505bf41832-Paper-Datasets_and_Benchmarks.pdf} {Judging llm-as-a-judge with mt-bench and chatbot arena}.
\newblock \emph{Advances in Neural Information Processing Systems}, 36.

\end{thebibliography}

\appendix
\section{Task Instructions}
\label{appendix:instructions}

We list the zero-shot instruction prompts we used for experiments in Section~\ref{sec:estimating} in \autoref{tab:instructions}.

\begin{table*}[ht]
\resizebox{\linewidth}{!}{
\begin{tabular}{@{}p{0.49\linewidth}p{0.49\linewidth}@{}}
\toprule
\textbf{Dataset} & \textbf{Instruction}                                                \\ \midrule
CNN/DailyMail \cite{nallapati-etal-2016-abstractive}    & Summarize the following article in 3-4 sentences.                   \\
SAMSum \cite{gliwa-etal-2019-samsum}           & Summarize the following dialogue in 1-2 sentences.                  \\
WMT-19 \cite{barrault-etal-2019-findings}          & Translate the following \{Czech, German\} sentence into English. \\ 
ART \cite{bhagavatula2019abductive}              & Write a hypothesis that explains the following observations.        \\
BiSECT \cite{kim-etal-2021-bisect}           & Write a simplification of the following sentence.  \\ 
\bottomrule                
\end{tabular}}
\caption{Prompt instructions for each benchmark used in \S\ref{sec:estimating}}
\label{tab:instructions}
\end{table*}

\begin{figure*}[ht]
     \begin{subfigure}[b]{0.49\textwidth}
         \centering
         \includegraphics[width=\textwidth]{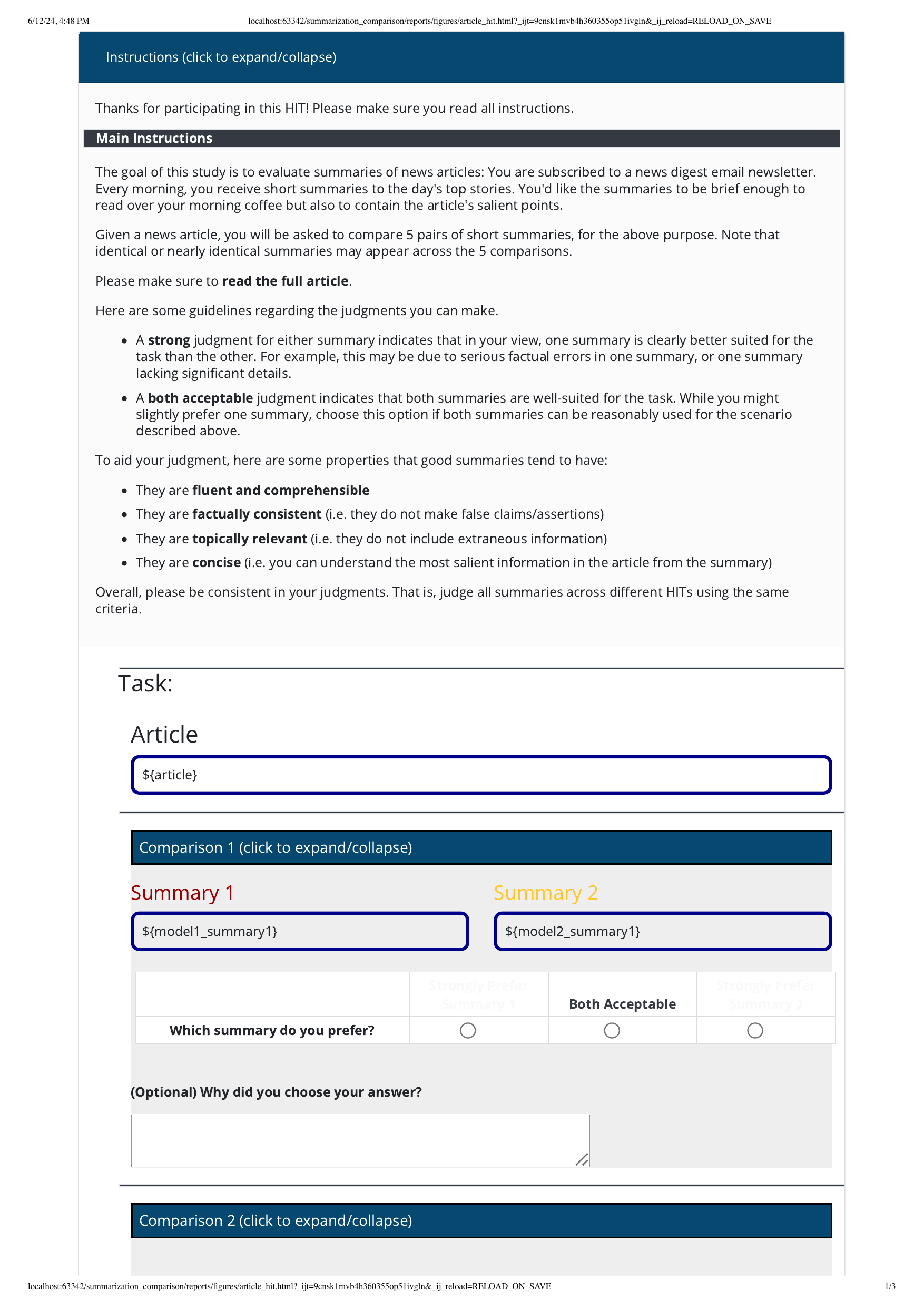}
     \end{subfigure}
     \begin{subfigure}[b]{0.49\textwidth}
         \centering
         \includegraphics[width=\textwidth]{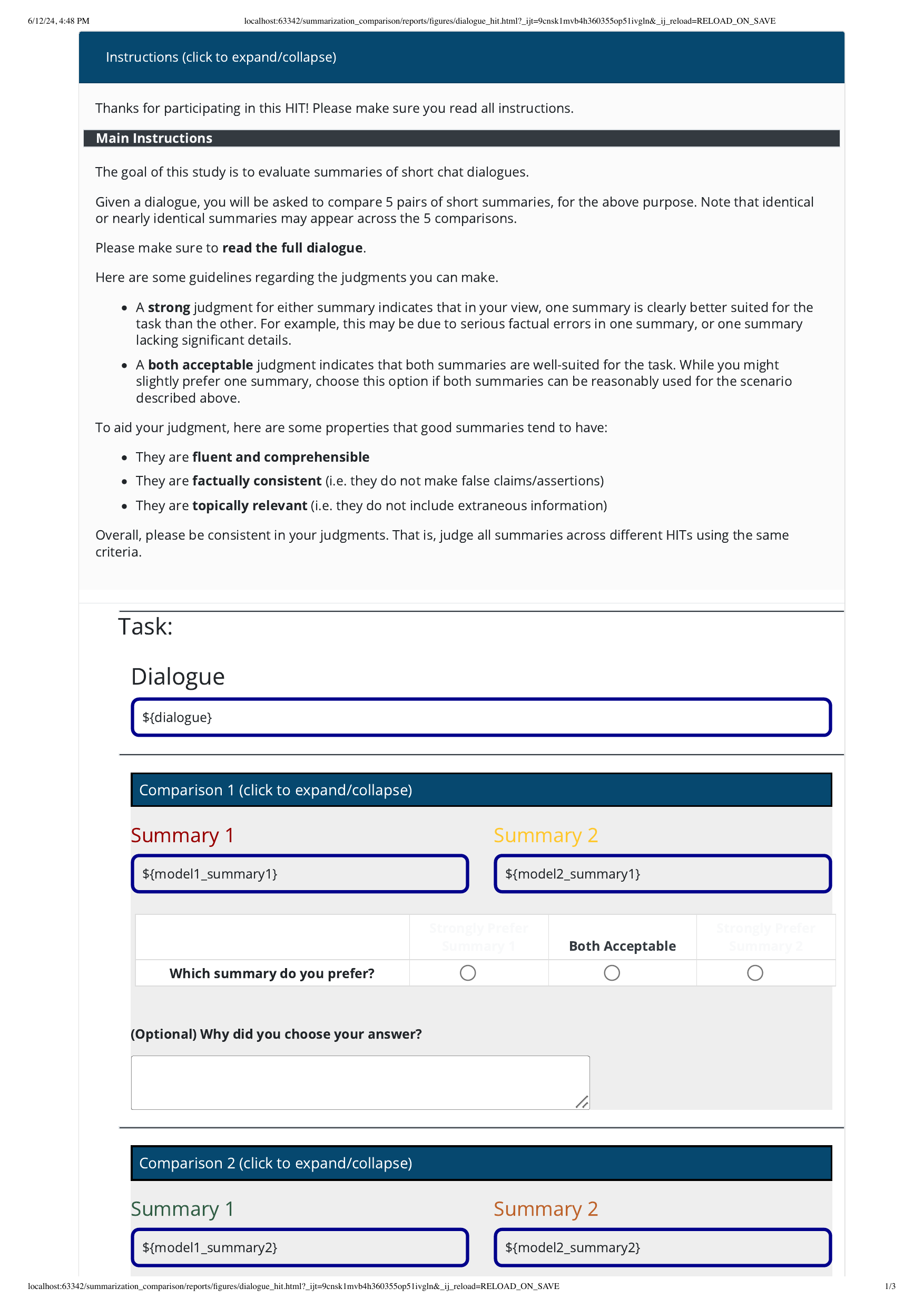}
     \end{subfigure}
        \caption{MTurk HIT Interface for news article (Left) and dialogue (Right) summary evaluation used in \S\ref{sec:human-study}}.
        \label{fig:hit-interface}
\end{figure*}

\section{Human Study Interface}
\label{appendix:hit-interface}

We present the HIT interface shown to AWS MTurk workers in Figure~\ref{fig:hit-interface}.

\section{Qualitative Examples}
\label{appendix:examples}
\begin{table*}[ht]
\small
\centering
\resizebox{\linewidth}{!}{
\begin{tabular}{@{}p{0.15\linewidth}p{0.42\linewidth}p{0.42\linewidth}}
\toprule
\textbf{}                & \multicolumn{2}{l}{\textbf{Lower Separability Instance Outputs}}   \\ \toprule
\textbf{Dataset}         & \textbf{Model A Generations} & \textbf{Model B Generations} \\ \toprule
CNN/DailyMail   & 
1. A Palestinian teenager's name was recently added to an Israeli memorial commemorating fallen soldiers and victims of terrorism, causing controversy and complaints from his family and an organization representing victims of terror. The teen's name, Mohammed Abu Khdeir, was ultimately removed from the memorial after his father's objections and a demand from the organization. The three suspects in Abu Khdeir's killing are currently on trial.

2. The name of a Palestinian teenager, Mohammed Abu Khdeir, was included on an Israeli memorial for fallen soldiers and victims of terrorism. His family and an organization representing victims of terror objected to his inclusion, and his name will now be removed from the memorial. The teenager was beaten and burned alive by three Israelis in July, and the suspects are currently on trial.

3. The name of a Palestinian teenager, Mohammed Abu Khdeir, was included on an Israeli memorial wall for fallen soldiers and victims of terrorism. His family and an organization representing victims of terror objected to his inclusion and demanded his name be removed. The National Insurance Institute of Israel has agreed to remove his name from the memorial. The teenager was beaten and burned alive by three Israelis in July and the suspects are currently going through a trial.                    
& 
A Palestinian teenager's name was added to an Israeli memorial for fallen soldiers and terrorism victims, but his family and a victim advocacy group objected, leading to the removal of his name. The teenager, Mohammed Abu Khdeir, was killed by three Israelis in July, and his family did not give permission for his name to be included. The National Insurance Institute of Israel will remove his name from the memorial following the complaints.

A Palestinian teenager's name was added to an Israeli memorial for fallen soldiers and terrorism victims, but his family and a victim advocacy group objected. The father of the teenager, who was killed in a brutal way, refused permission for his son's name to be listed among soldiers of the occupation. The National Insurance Institute of Israel will remove the teenager's name from the memorial following the complaints. The teenager's name has already been removed from the organization's website.

A Palestinian teenager's name was added to an Israeli memorial for fallen soldiers and terrorism victims, but his family and a victim advocacy group objected. The father of the teenager, who was killed in a brutal way, said his son's name should not be listed among soldiers of the occupation. The National Insurance Institute of Israel, which maintains the memorial site, will remove the teenager's name following the complaints. The judicial hearing for the suspects in the teenager's killing began in January.             
\\ \midrule
SAMSum & 
1. Ursula shares her success of getting a 93 on her French exam, while her friends congratulate her and discuss their own experiences with the exam.

2. Ursula shares her achievement of getting a 93 on her French exam, and her friends congratulate her while discussing their own experiences with the exam.

3. Ursula shares her success of getting a 93 on her French exam, while Vaughn admits to getting a 65 without studying. Jana and Bob congratulate Ursula while Vaughn jokes about his own score.

%4. Ursula shares her success of getting a 93 on her French exam, and her friends congratulate her while discussing their own experiences with the subject.

%5. Ursula got a high grade on her French exam and credits her hard work for her success. Vaughn, on the other hand, did not study and got a lower grade.
& 
1. Ursula got a 93 on her French exam, while Vaughn got a 65. Jana is impressed with Ursula's score and asks how she managed it, to which Ursula replies that she studied hard. Vaughn admits that he didn't study and only got a passing grade.

2. Ursula got a 93 on her French exam, Bob congratulated her, Jana was impressed, and Vaughn got a 65 but didn't study for it.

3. Ursula got a 93 on her French exam, Bob congratulated her, Jana was impressed, and Vaughn got a 65 but didn't study for it.

%4. Ursula got a 93 on her French exam, while Vaughn got a 65. Ursula studied hard for the exam, while Vaughn did not.

%5. Ursula got a 93 on her French exam, Bob congratulated her, Jana was impressed, and Vaughn got a 65 but didn't study for it.
\\ \bottomrule
\end{tabular}}
\caption{Examples of lower \separability instances for summarization benchmarks used in our experiments}
\label{tab:qual-examples-low-summ}
\end{table*}

\begin{table*}[ht]
\small
\centering
\resizebox{\linewidth}{!}{
\begin{tabular}{@{}p{0.15\linewidth}p{0.42\linewidth}p{0.42\linewidth}}
\toprule
\textbf{}                & \multicolumn{2}{l}{\textbf{Higher Separability Instance Outputs}}   \\ \toprule
\textbf{Dataset}         & \textbf{Model A Generations} & \textbf{Model B Generations} \\ \toprule
CNN/DailyMail   & 
1. Ciudad Juarez, Mexico, which was once known as the murder capital of the world with an average of 8.5 killings per day in 2010, is now making efforts to attract foreign tourists and investors back to the city. The city launched the tourism campaign "Juarez is Waiting for You" this month. Violence in the city has decreased significantly, with only 89 killings so far this year compared to over 3,000 in 2014. 

2. Ciudad Juarez, Mexico, which was once known as the murder capital of the world, is now making efforts to lure tourists and investors back with a new tourism campaign called "Juarez is Waiting for You." In 2010, the city averaged 8.5 killings per day, but local officials claim that it is much safer now. The violent past, which included a turf war between cartels and high levels of corruption within the local police force, has left deep scars. 

3. Ciudad Juarez, Mexico, which was once known as the murder capital of the world with an average of 8.5 killings per day in 2010, is now making efforts to lure foreign tourists and investors back. The city launched the tourism campaign "Juarez is Waiting for You" this month, and local officials claim that the city is much safer than it was five years ago. In 2014, there were 424 homicides, but this year, there have been 89 killings, a significant decrease in violence.                   
& 
1. Ciudad Juarez, Mexico, was once the murder capital of the world, with an average of 8.5 killings per day in 2010. However, local officials say the city is much safer now, and plans are underway to lure foreign tourists and investors back to Juarez.

2. Ciudad Juarez, Mexico, was once the murder capital of the world, with an average of 8.5 killings per day in 2010. However, local officials say the city is much safer now, and plans are underway to lure foreign tourists and investors back to Juarez.

3. Ciudad Juarez, Mexico, was once the murder capital of the world, with an average of 8.5 killings per day in 2010. However, local officials say the city is much safer now, and plans are underway to lure foreign tourists and investors back to Juarez.
\\ \midrule
SAMSum & 
1. Tom agrees to pick up Susan's daughter Becky from school because she has a fever and Susan cannot leave the office. Tom also needs to return to the office soon, so he will take Becky to his mother's house.

2. Tom agrees to pick up Susan's daughter Becky from school because she has a fever and Susan cannot leave the office. Tom also needs to return to the office to finish a project.

3. Tom will pick up Susan's daughter Becky from school because she has a fever and Susan cannot leave the office. Tom also needs to return to the office to finish a project.
& 
1. Susan's day was disappointing due to a failed presentation, and she asked Tom to pick up their sick daughter Becky from school since she couldn't leave work early. Tom agreed and planned to take Becky to his mother's house. Susan expressed her frustration over a disappointing day, including a failed presentation, and requested Tom's help in picking up their sick daughter Becky from school due to her inability to leave work early. Tom agreed and shared his own work pressure, intending to take Becky to his mother's house.

2. Susan shares that her day was disappointing due to a failed presentation, and reveals that her daughter Becky is sick at school, requiring Tom to pick her up and take her to his mother's house, as Susan is unable to leave work early and doesn't have her car. Tom mentions the pressure he's under to complete a project on time.

3. Susan's day was disappointing due to a failed presentation, and she asked Tom to pick up their sick daughter Becky from school as she couldn't leave work early and didn't have her car. Tom agreed, mentioning his own work pressure. Susan had a frustrating day with her presentation going wrong, and she asked Tom to pick up their sick daughter Becky from school as she couldn't leave work early and didn't have her car; Tom agreed, mentioning his own work pressure and offering to take Becky to his mother's house.
\\ \bottomrule
\end{tabular}}
\caption{Examples of higher \separability instances for summarization benchmarks used in our experiments}
\label{tab:qual-examples-high-summ}
\end{table*}

\begin{table*}[ht]
\small
\centering
\resizebox{\linewidth}{!}{
\begin{tabular}{@{}p{0.1\linewidth}p{0.44\linewidth}p{0.44\linewidth}}
\toprule
\textbf{}                & \multicolumn{2}{l}{\textbf{Lower Separability Instance Outputs}}   \\ \toprule
\textbf{Dataset}         & \textbf{Model A Generations} & \textbf{Model B Generations} \\ \toprule
ART   & 
1. Regular practice and exposure to rock climbing with friends can improve an individual's climbing ability and confidence, leading to better performance on a real rock wall.

2. My hypothesis is that the practice and experience gained from indoor rock climbing with friends improved my skills and confidence, leading to better performance on a real rock wall the following day.

3. Participating in indoor rock climbing with friends can improve an individual's climbing abilities and confidence, leading to better performance on a real outdoor rock wall.                   
& 
1. The experience of rock climbing with friends and the subsequent visit to the real rock wall helped me develop a better understanding of the techniques and skills required for rock climbing, which improved my performance on the real rock wall.

2. The experience of rock climbing with friends, even if it was just pretend, helped improve my skills and confidence in real rock climbing.

3. The experience of rock climbing with friends, even if it was just pretend, helped me develop the skills and confidence needed to successfully climb a real rock wall.

\\ \midrule
BiSECT & 
1. Stiripentol is absorbed quickly, reaching its peak concentration in about 1.5 hours.

2. Stiripentol is absorbed quickly, reaching its peak concentration in about 1.5 hours.

3. Stiripentol is rapidly absorbed, reaching its peak plasma concentration in 1.5 hours.
& 
1. Stiripentol is rapidly absorbed, reaching its peak concentration in approximately 1.5 hours.

2. Stiripentol is rapidly absorbed, reaching its peak concentration in approximately 1.5 hours.

3. Stiripentol is rapidly absorbed, reaching its peak concentration in approximately 1.5 hours.
\\ \bottomrule
\end{tabular}}
\caption{Examples of lower \separability instances for ART and BiSECT}
\label{tab:qual-examples-low-other}
\end{table*}

\begin{table*}[ht]
\small
\centering
\resizebox{\linewidth}{!}{
\begin{tabular}{@{}p{0.1\linewidth}p{0.44\linewidth}p{0.44\linewidth}}
\toprule
\textbf{}                & \multicolumn{2}{l}{\textbf{Higher Separability Instance Outputs}}   \\ \toprule
\textbf{Dataset}         & \textbf{Model A Generations} & \textbf{Model B Generations} \\ \toprule
ART   & 
1. Joey's desire to color in coloring books and his wish for his parents to appreciate him may be related to his need for attention and validation.

2. Joey's desire to color in coloring books and his wish for his parents to appreciate him may be related to his need for attention and validation.

3. Joey's desire to color in coloring books and his wish for his parents to appreciate him may be related to his need for attention and validation.                   
& 
1. Hypothesis: Joey's engagement in coloring activities in his coloring books could be an expression of his innate need for creativity and self-expression, which may serve as a coping mechanism to deal with his unmet emotional need for appreciation from his parents. 2. Explanation: Joey's enjoyment of coloring might provide him with a sense of accomplishment, relaxation, and control, which can help him feel good about himself.

2. Hypothesis: Joey's engagement in coloring as a second-grader could be an expression of his desire for approval and appreciation from his parents. The act of coloring may provide him with a sense of accomplishment, creativity, and relaxation, which in turn, could be a way for him to express his emotions and seek validation from his caregivers. However, the fact that he expressed a wish for his parents to appreciate him more suggests that he may not feel fully valued or recognized for his efforts and talents outside of this activity.

3. Hypothesis: Joey's engagement in coloring activities in his coloring books could be an expression of his desire for recognition and appreciation from his parents. This hypothesis is based on the observation that Joey, as a second-grader, enjoys coloring, but also expressed a wish for his parents to value and acknowledge him. It's possible that the act of coloring provides Joey with a sense of accomplishment and pride, and he may hope that his parents will notice and praise him for his efforts.

\\ \midrule
BiSECT & 
1. The rise in cheap imports harms Community producers by putting downward pressure on prices and decreasing their sales.

2. The rise in cheap imports harms Community producers by putting pressure on prices and decreasing their sales volume in the market.

3. The increase in low-priced imports is harming Community producers by putting pressure on prices and decreasing their sales volume in the Community market.
& 
1. There is a correlation between the rise in imports at low prices and the harm suffered by European producers. The increase in imports has had negative effects on prices and sales volume for European producers in the European market.

2. There is a connection between the rise in imports at low prices and the harm suffered by European producers. This increase in imports has negatively impacted prices and reduced the volume sold by European producers in the European market.

3. There is a connection between the rise in low-priced imports and the harm suffered by European producers. This increase in imports has caused negative effects, such as pushing down prices and reducing the amount of goods sold by European producers in their home market.
\\ \bottomrule
\end{tabular}}
\caption{Examples of higher \separability instances for ART and BiSECT}
\label{tab:qual-examples-high-other}
\end{table*}

\begin{table*}[ht]
\small
\centering
\resizebox{\linewidth}{!}{
\begin{tabular}{@{}p{0.1\linewidth}p{0.44\linewidth}p{0.44\linewidth}}
\toprule
\textbf{}                & \multicolumn{2}{l}{\textbf{Lower Separability Instance Outputs}}   \\ \toprule
\textbf{Dataset}         & \textbf{Model A Generations} & \textbf{Model B Generations} \\ \toprule
WMT-19 DE to EN   & 
1. If the offer arrives, we will expand it.

2. If the offer arrives, we will expand it.

3. If the offer arrives, we will expand it.
                
& 
1. If the offer comes, we will expand it.

2. If the offer comes, we'll expand it.

3. If the offer comes, we'll expand it.

\\ \midrule
WMT19 CS to EN & 
1. Papuans rarely have the necessary capital and education to start a business.

2. Papuans rarely have the necessary capital and education to start a business.

3. Papuans rarely have the necessary capital and education to start a business.
& 
1. Papuans rarely have the necessary capital and education to start a business.

2. Papuans rarely have the necessary capital and education to start a business.

3. Papuans rarely have the necessary capital and education to start a business.
\\ \bottomrule
\end{tabular}}
\caption{Examples of lower \separability instances for translation benchmarks used in our experiments}
\label{tab:qual-examples-low-trans}
\end{table*}

\begin{table*}[ht]
\small
\centering
\resizebox{\linewidth}{!}{
\begin{tabular}{@{}p{0.1\linewidth}p{0.44\linewidth}p{0.44\linewidth}}
\toprule
\textbf{}                & \multicolumn{2}{l}{\textbf{Higher Separability Instance Outputs}}   \\ \toprule
\textbf{Dataset}         & \textbf{Model A Generations} & \textbf{Model B Generations} \\ \toprule
WMT-19 DE to EN   & 
1. In 2013, a total of 25 plots were fenced in by eleven participants through the pasture fence promotion.

2. In 2013, a total of 25 plots were fenced in by eleven participants through the pasture fence promotion.

3. In 2013, a total of 25 plots were fenced in by eleven participants through the pasture fence promotion.
                
& 
1. 11 participants enclosed 25 parcels with fencing in 2013.

2. 11 participants enclosed 25 parcels with fencing in 2013.

3. 11 participants enclosed 25 parcels through the pasture fencing promotion in the year 2013.

\\ \midrule
WMT19 CS to EN & 
1. They shoot less, and even when they have a half-empty goal in front of them, they prefer to pass.

2. They shoot less, and even when they have a half-empty goal in front of them, they prefer to pass.

3. They shoot less, and even when they have a half-empty goal in front of them, they prefer to pass.
& 
1. I prefer to shoot when there is an empty space in front of me, even if it means taking a risk.

2. I prefer to shoot when there is an empty space in front of me, even if it means taking a risk.

3. I prefer to shoot when there is an empty space in front of me, rather than when I have an empty space behind me.
\\ \bottomrule
\end{tabular}}
\caption{Examples of higher \separability instances for translation benchmarks used in our experiments}
\label{tab:qual-examples-high-trans}
\end{table*}
We present examples of generations corresponding to higher and lower \separability instances for the benchmarks used in our experiments in Tables~\ref{tab:qual-examples-low-summ} to ~\ref{tab:qual-examples-high-trans}.

\section{Preference Strength}
\label{appendix:strength-pref}

\begin{figure*}[h]
     \centering
     \includegraphics[width=\textwidth]{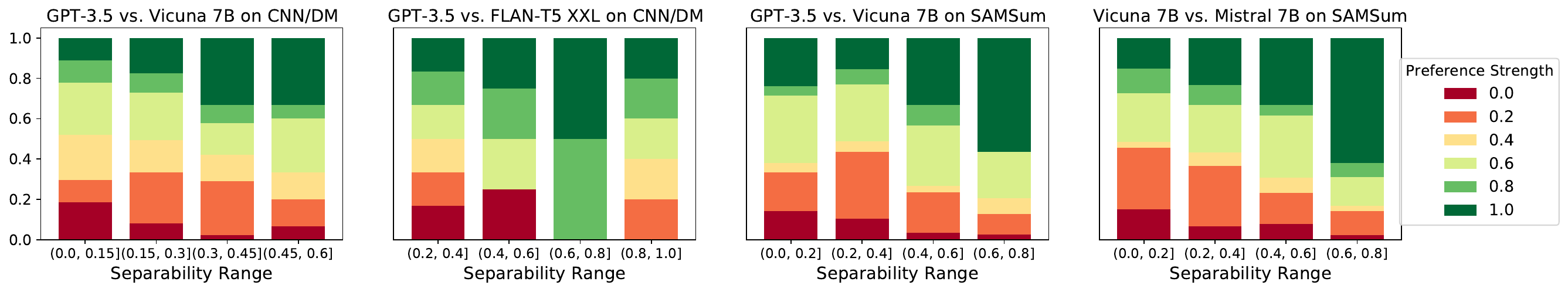}
     \vspace*{-6mm}
            \caption{Proportion of instances with each possible preference strength value in a \separability range, with human raters}
        \label{fig:strength_stacks}
\end{figure*}

\begin{figure*}[h]
     \centering
     \includegraphics[width=\textwidth]{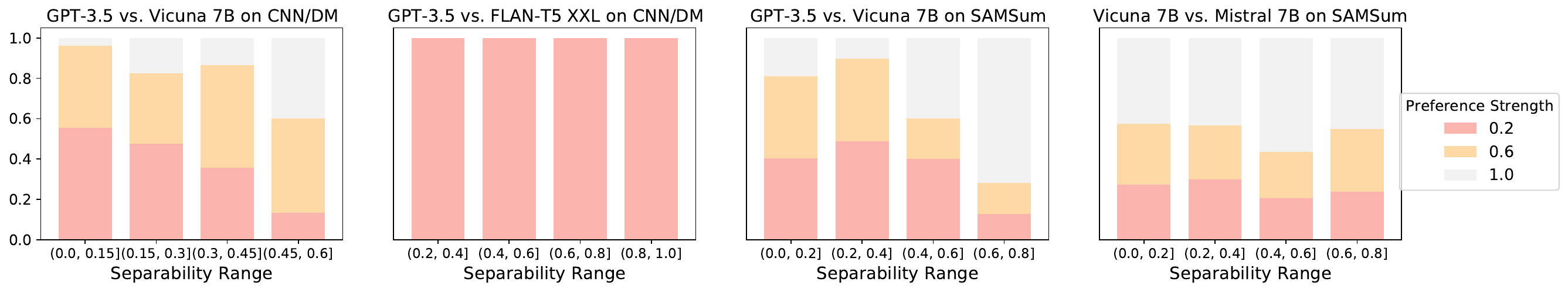}
     \vspace*{-6mm}
            \caption{Proportion of instances with each possible preference strength value in a \separability range, with auto-raters.}
        \label{fig:strength_stacks_alpaca}
\end{figure*}

In addition to consistency (\autoref{eq:consistency}), we define define another way to determine how much a rating set shows preference towards one model or another. 
We call this metric \textbf{preference strength}.
We recycle the notation from \autoref{eq:consistency} and define \textbf{preference strength} of a rating set $\mathcal{R}_{a}(\textbf{x}^i)$ as:
\begin{equation}
\resizebox{\columnwidth}{!}{
$\textsf{pref-strength}(\mathcal{R}_a(\mathbf{x}^i)) = \frac{\sum_{j=1}^N r_a\left(\left(\mathbf{y}_A^{i, j}, \mathbf{y}_B^{i, j}\right)\right)}{N}$}
\end{equation}
Intuitively, preference strength is simply the mean of the rating set. 
Preference strength of $-1, 1$ reflects a rating set where all the ratings were towards model $A, B$ respectively.

We present the proportion of instances with each possible preference strength per \separability range in Figures~\ref{fig:strength_stacks} and Figures~\ref{fig:strength_stacks_alpaca}.

\section{Using other similarity metrics}
\label{appendix:alt-metrics}
In Section~\ref{sec:sim-fcns}, we describe how different similarity functions can be used for different tasks, as well as to measure different dimensions of \separability.

\begin{figure*}[h]
     \centering
     \includegraphics[width=\textwidth]{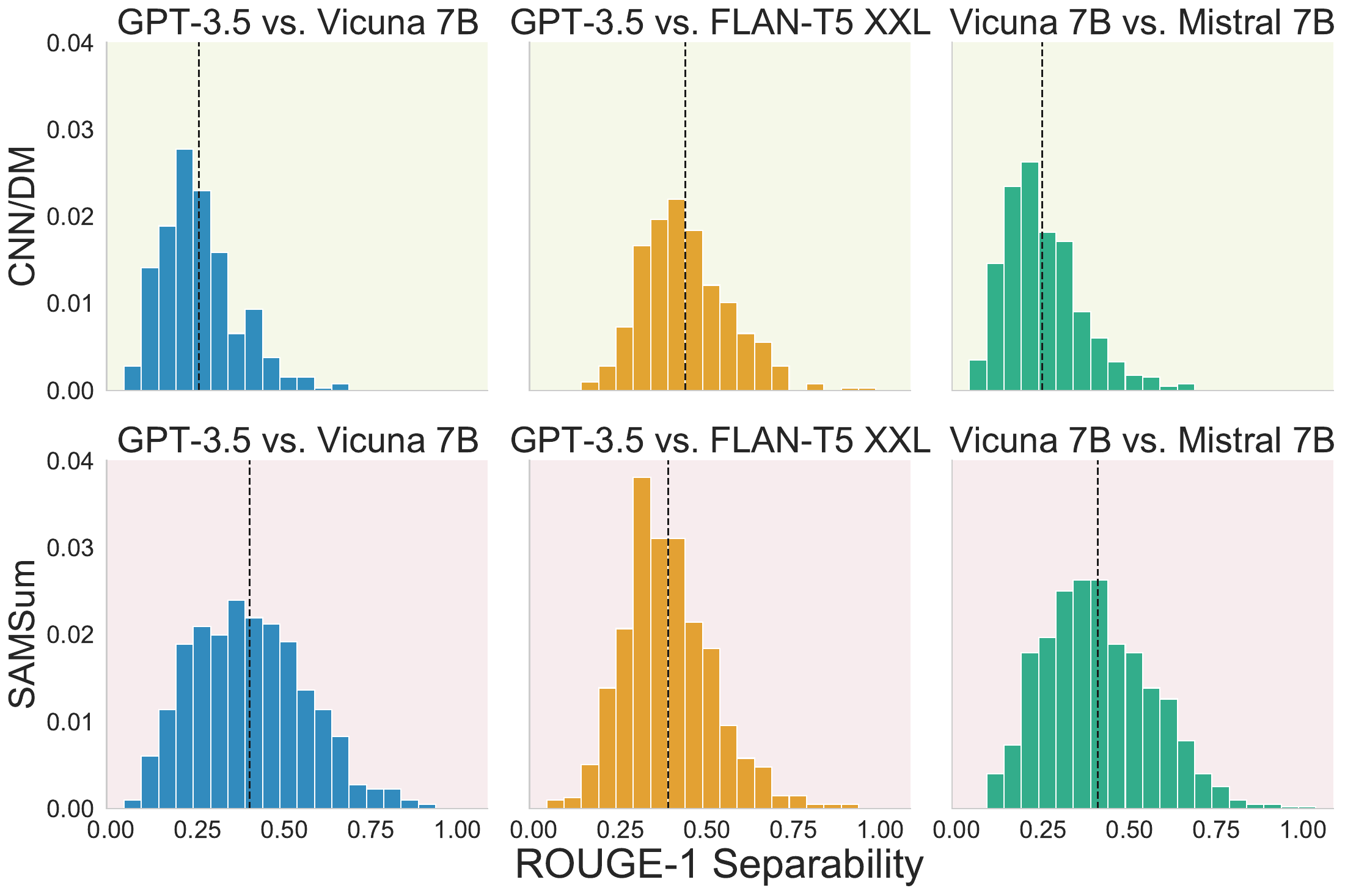}
     \vspace*{-6mm}
            \caption{\separability distributions using ROUGE-1 F1 as a similarity metric. }
        \label{fig:sep-hist-rouge}
\end{figure*}

\begin{figure*}[h]
     \centering
     \includegraphics[width=\textwidth]{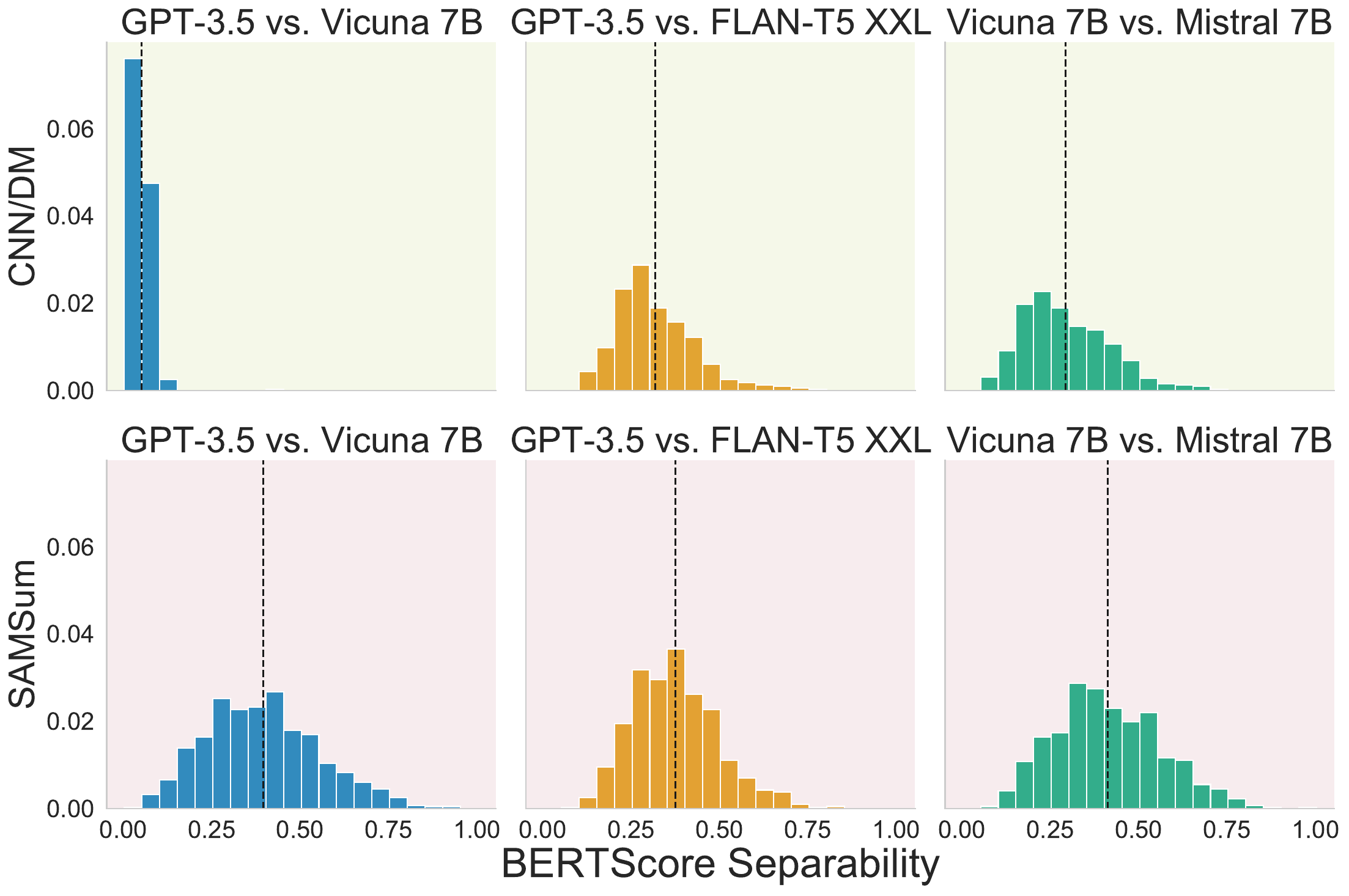}
     \vspace*{-6mm}
            \caption{\separability distributions using vanilla BERTScore as a similarity metric. }
        \label{fig:sep-hist-bertscore}
\end{figure*}

\begin{figure*}[h]
     \centering
     \includegraphics[width=\textwidth]{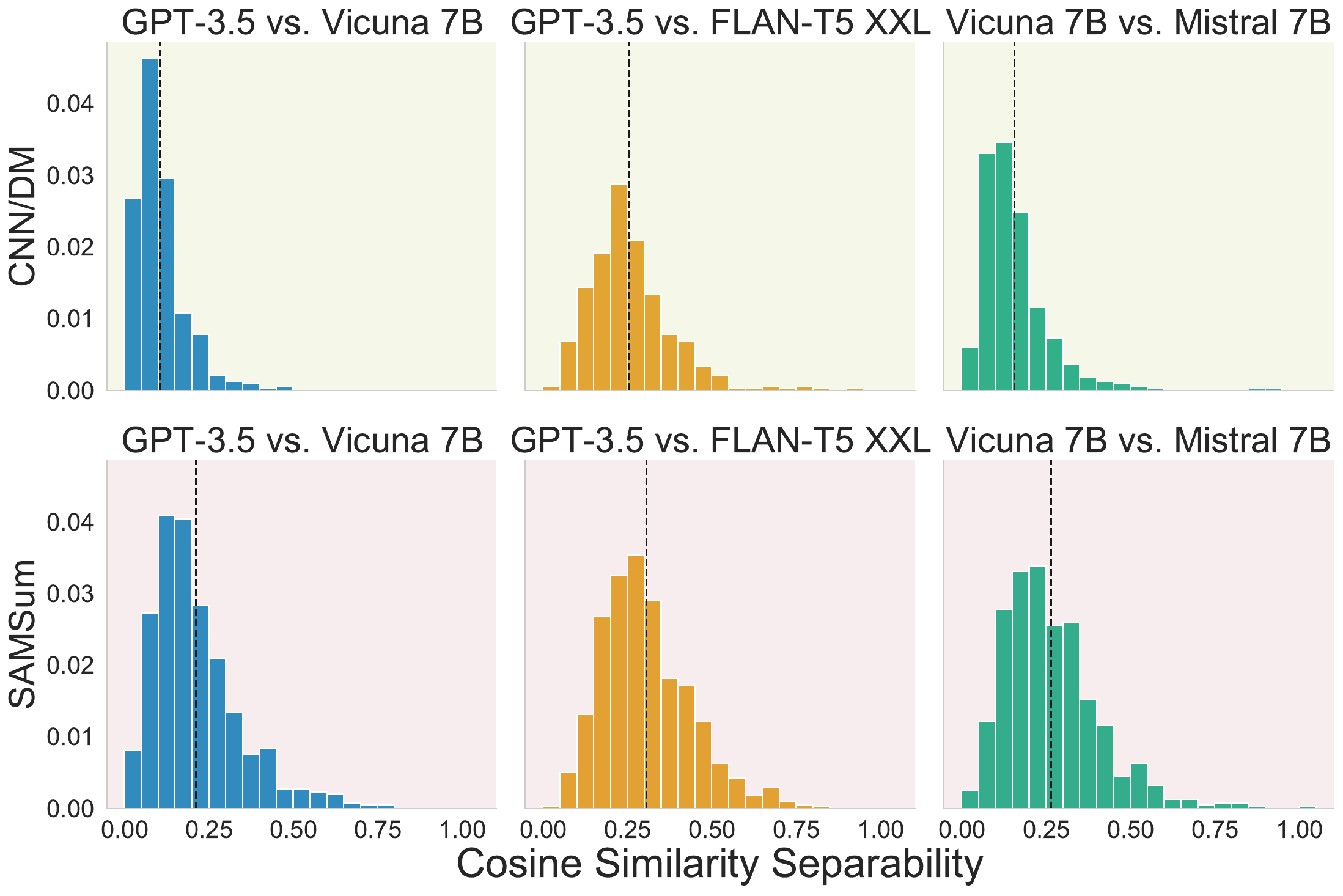}
     \vspace*{-6mm}
            \caption{\separability distributions using cosine similarity as a similarity metric. }
        \label{fig:sep-hist-cosine}
\end{figure*}

\begin{figure*}[h]
     \centering
     \includegraphics[width=\textwidth]{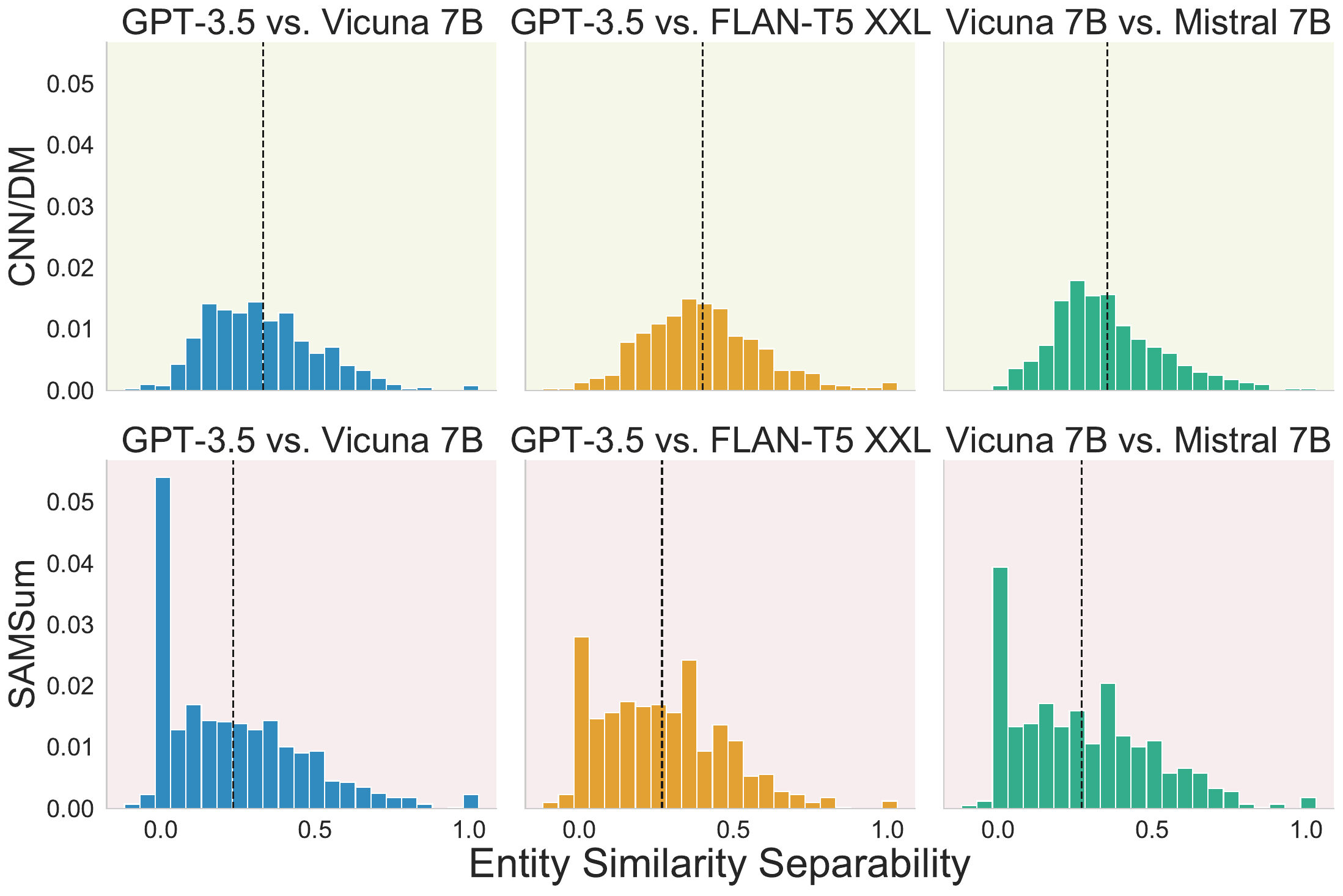}
     \vspace*{-6mm}
            \caption{\separability distributions using entity similarity as a similarity metric.}
        \label{fig:sep-hist-entity}
\end{figure*}

We present \separability distributions in Figures~\ref{fig:sep-hist-rouge} to \ref{fig:sep-hist-entity} for ROUGE-1 \citep{lin-2004-rouge}, the original BERTScore \citep{zhangbertscore}, entity similarity\footnote{We calculate entity similarity between two generations by using \texttt{spacy} to extract named entities and taking the Jaccard Similarity of the set of entities from each generation}, and cosine similarity\footnote{Using \texttt{all-distilroberta-v1} in the \texttt{sentence-transformers} library}, and text embedding cosine similarity on the summarization benchmarks used in our experiments.

\end{document}